\documentclass[conference]{IEEEtran}
\ifCLASSINFOpdf
   \usepackage[pdftex]{graphicx}
   \DeclareGraphicsExtensions{.pdf,.jpeg,.png}
\else
\fi
\newcommand{\chris}[1]{{\footnotesize\color{red}[Chris: #1]}}
\newcommand{\kartik}[1]{{\footnotesize\color{green}[Kartik: #1]}}
\newcommand{\jiyong}[1]{{\footnotesize\color{blue}[Jiyong: #1]}}

\newcommand{\inp}{\mathsf{I}}
\newcommand{\out}{\mathsf{O}}
\newcommand{\filt}{\mathsf{F}}

\newcommand{\indtable}{\mathsf{iiT}}
\newcommand{\wttable}{\mathsf{wiT}}

\newcommand{\conf}[1]{$\mathsf{#1}$}

\newcommand{\para}[1]{\vspace{.05in} \noindent \textbf{#1}}

\usepackage{mathptmx} 

\newcommand{\ignore}[1]{}
\usepackage{fancyhdr}
\usepackage{microtype}
\usepackage[normalem]{ulem}
\usepackage[hyphens]{url}
\usepackage{color}
\usepackage{comment}
\usepackage{paralist}
\usepackage[]{algorithm2e}
\usepackage{pifont}
\usepackage{soul}
\usepackage{listings}
\usepackage{graphicx}
\usepackage{mathtools}
\usepackage{changepage}

\makeatletter
\def\verbatim{\small\@verbatim \frenchspacing\@vobeyspaces \@xverbatim}
\makeatother
\usepackage[bookmarks=true,breaklinks=true,letterpaper=true,colorlinks,linkcolor=blue,citecolor=green,urlcolor=blue]{hyperref}

\definecolor{dkgreen}{rgb}{0,0.6,0}
\definecolor{gray}{rgb}{0.5,0.5,0.5}
\definecolor{mauve}{rgb}{0.58,0,0.82}

\begin{document}

%
\title{UCNN: Exploiting Computational Reuse in Deep Neural Networks via Weight Repetition}
\author{Kartik Hegde, 
        Jiyong Yu, 
        Rohit Agrawal, 
        Mengjia Yan,
        Michael Pellauer$^*$, 
        Christopher W. Fletcher\\
        University of Illinois at Urbana-Champaign \\ $^*$NVIDIA \\
        \{kvhegde2, jiyongy2, rohita2, myan8\}@illinois.edu, mpellauer@nvidia.com, cwfletch@illinois.edu}

\maketitle
\thispagestyle{fancy}
\chead{Appears in the proceedings of the 45th International Symposium on Computer Architecture~(ISCA), 2018}

\begin{abstract}

Convolutional Neural Networks (CNNs) have begun to permeate all corners of electronic society (from voice recognition to scene generation) due to their high accuracy and machine efficiency per operation.
At their core, CNN computations are made up of multi-dimensional dot products between weight and input vectors.
This paper studies how \emph{weight repetition}---when the same weight occurs multiple times in or across weight vectors---can be exploited to save energy and improve performance during CNN inference.
This generalizes a popular line of work to improve efficiency from CNN weight sparsity, as reducing computation due to repeated zero weights is a special case of reducing computation due to repeated weights.

To exploit weight repetition, this paper proposes a new CNN accelerator called the \textbf{U}nique Weight \textbf{CNN} Accelerator (UCNN).
UCNN uses weight repetition to reuse CNN sub-computations (e.g., dot products) and to reduce CNN model size when stored in off-chip DRAM---both of which save energy.
UCNN further improves performance by exploiting sparsity in weights.
We evaluate UCNN with an accelerator-level cycle and energy model and with an RTL implementation of the UCNN processing element.
On three contemporary CNNs, UCNN improves throughput-normalized energy consumption by $1.2\times\sim4\times$, relative to a similarly provisioned baseline accelerator that uses Eyeriss-style sparsity optimizations.
At the same time, the UCNN processing element adds only 17-24\% area overhead relative to the same baseline.




\end{abstract}

\section{Introduction}
\label{sec:intro}

%
%

We are witnessing an explosion in the use of Deep Neural Networks (DNNs), with major impacts on
the world's economic and social activity. 
At present, there is abundant evidence of DNN's effectiveness in areas such as classification, vision, and speech~\cite{DNN_Speech,DNN_Image,DNN_Economy,DNN_Medicine,DNN_Bio}.
Of particular interest are Convolutional Neural Networks (CNNs), which achieve state-of-the-art performance in many of these areas, such as image/temporal action recognition~\cite{Alexnet,TwoStreamConv} and scene generation~\cite{CNNGan}.
An ongoing challenge is to bring CNN \emph{inference}---where the CNN is deployed in the field and asked to answer online queries---to edge devices, which has inspired CNN architectures ranging from CPUs to GPUs to custom accelerators~\cite{DaDianNao,Cnvlutin,EIE,SCNN}.
A major challenge along this line is that CNNs are notoriously compute intensive~\cite{DaDianNao,TPU}.
It is imperative to find new ways to reduce the work needed to perform inference.



%
%
%

\begin{figure}[ht!]
  \begin{centering}
  \includegraphics[width=\columnwidth]{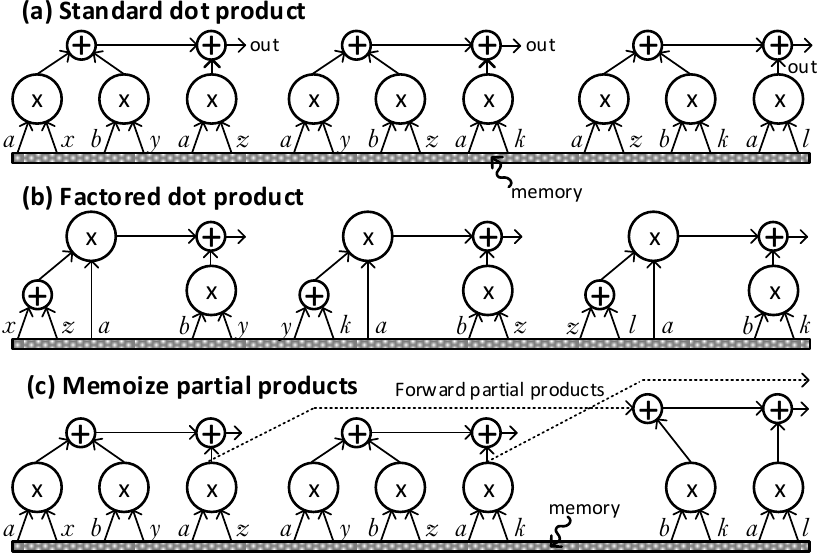}
  \caption{\label{fig:schemes}
            \small 
            Standard (a) and different optimized (b, c) 1D convolutions that take advantage of repeated weight $a$.  
            Arrows out of the grey bars indicate input/filter memory reads.
            Our goal is to reduce memory reads, multiplications and additions while obtaining the same result.
            } 
  \end{centering}
\end{figure}

%
%
At their core, CNN computations are parallel dot products.
Consider a 1-dimensional convolution (i.e., a simplified CNN kernel), which has filter $\{a, b, a\}$ and input $\{x, y, z, k, l, \dots\}$.
We refer to elements in the input as the activations, elements in the filter as the weights and the number of weights in the filter as the filter's size (3 in this case).
The output is computed by sliding the filter across the input and taking a filter-sized dot product at each position (i.e., $\{ax + by + az, ay + bz + ak, \dots\}$), as shown in Figure~\ref{fig:schemes}a.
When evaluated on hardware, this computation entails reading each input and weight from memory, and performing a multiply-accumulate (MAC) on that input-weight pair.
In this case, each output performs 6 memory reads (3 weights and 3 inputs), 3 multiplies and 2 adds.

%
%
In this paper, we explore how CNN hardware accelerators can eliminate superfluous computation by taking advantage of \emph{repeated weights}.
In the above example, we have several such opportunities because the weight $a$ appears twice.
First (Figure~\ref{fig:schemes}b), we can factor dot products as sum-of-products-of-sums expressions, saving 33\% multiplies and 16\% memory reads.
Second (Figure~\ref{fig:schemes}c), each partial product $a*\mathrm{input}$ computed at the filter's right-most position can be memoized and re-used when the filter slides right by two positions, saving 33\% multiples and memory reads.
Additional opportunities are explained in Section~\ref{sec:comp_reuse}.
Our architecture is built on these two ideas: factorization and memoization, both of which are only possible given repeated weights (two $a$'s in this case).
Reducing computation via weight repetition is possible due to CNN filter design/weight quantization techniques, and is inspired by recent work on sparse CNN accelerators. 
A filter is guaranteed to have repeated weights when the filter size exceeds the number of unique weights,  
due to the pigeonhole principle.
Thus, out-of-the-box (i.e., not re-trained~\cite{DeepCompression}) networks may see weight repetition already.
For example, representing each weight in 8 bits~\cite{TPU} implies there are $\le 256$ unique weights, whereas filter size can be in the thousands of weights~\cite{Alexnet,Googlenet,Resnet}. 
Augmenting this further is a rich line of work to \emph{quantize weights}~\cite{DeepCompression,Ternary,INQ}, which strives to decrease the number of unique weights without losing significant classification accuracy.
For example, INQ~\cite{INQ} and TTQ~\cite{Ternary} use 17 and 3 unique weights, respectively, without changing filter size.
Finally, innovations~\cite{Cnvlutin,Cambriconx,SCNN} that exploit CNN sparsity (zero weights/activations) inspire and complement weight repetition.  
Weight repetition generalizes this optimization: reducing computation due to repeated zero weights is a special case of reducing computation due to repeated weights.


%
%
Exploiting weight repetition while getting a net efficiency win, however, is challenging for two reasons.
First, as with sparse architectures, tracking repetition patterns is difficult because they are irregular.
Second, na\"ive representations of tracking metadata require a large amount of storage. 
This is a serious problem due to added system energy cost of transporting metadata throughout the system (e.g., reading the model from DRAM, PCI-e, etc). 

This paper addresses these challenges with a novel CNN accelerator architecture called UCNN, for \underline{U}nique Weight \underline{CNN} Accelerator.
UCNN is based on two main ideas.
First, we propose a \emph{factorized dot product dataflow} which reduces multiplies and weight memory reads via weight repetition, and improves performance via weight sparsity~\cite{Cambriconx, SCNN}.
Second, we propose \emph{activation group reuse}, which builds on dot product factorization to reduce input memory reads, weight memory reads, multiplies and adds per dot product, while simultaneously compressing the CNN model size.
The compression rate is competitive to that given by aggressive weight quantization schemes~\cite{INQ,Ternary}, and gives an added ability to exploit weight repetition.
We employ additional architectural techniques to amortize the energy cost of irregular accesses and to reduce hardware area overhead.

\para{Contributions.}
To summarize, this paper makes the following contributions.
\begin{compactenum}
    \item We introduce new techniques---including dot product factorization and activation group reuse---to improve CNN efficiency by exploiting weight repetition in and across CNN filters.
    \item We design and implement a novel CNN accelerator, called UCNN, that improves performance and efficiency per dot product by using the aforementioned techniques.
    \item We evaluate UCNN using an accelerator-level cycle and energy model as well as an RTL prototype of the UCNN processing element.
On three contemporary CNNs, UCNN improves throughput-normalized energy consumption by $1.2\times\sim4\times$, relative to a similarly provisioned baseline accelerator that uses Eyeriss-style sparsity optimizations.
At the same time, the UCNN processing element adds only 17-24\% area overhead relative to the same baseline.

\end{compactenum}

We note that while our focus is to accelerate CNNs due to their central role in many problems, weight repetition is a general phenomena that can be exploited by any DNN based on dot products, e.g., multilayer perceptions.
Further, some of our techniques, e.g., dot product factorization, work out of the box for non-CNN algorithms.

\para{Paper outline.}
The rest of the paper is organized as follows.
Section~\ref{sec:background} gives background on CNNs and where weight repetition occurs in modern networks.
Section~\ref{sec:comp_reuse} presents strategies for CNN accelerators to reduce work via weight repetition.
Section~\ref{sec:architecture} proposes a detailed processing element (PE)-level architecture to improve efficiency via weight repetition.
Section~\ref{sec:dataflow} gives a dataflow and macro-architecture for the PE.
Section~\ref{sec:evaluation} evaluates our architecture relative to dense baselines.
Section~\ref{sec:related} covers related work.
Finally Section~\ref{sec:conclusion} concludes.

\begin{figure}[]
  \begin{centering}
  \includegraphics[width=.9\columnwidth]{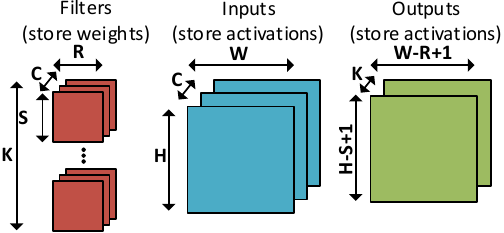}
  \caption{\label{fig:conv_params}
            \small 
            CNN parameters per convolutional layer.} 
  \end{centering}
\end{figure}

\section{Background}
\label{sec:background}

\begin{figure*}[t]
  \begin{centering}
  \includegraphics[width=\textwidth]{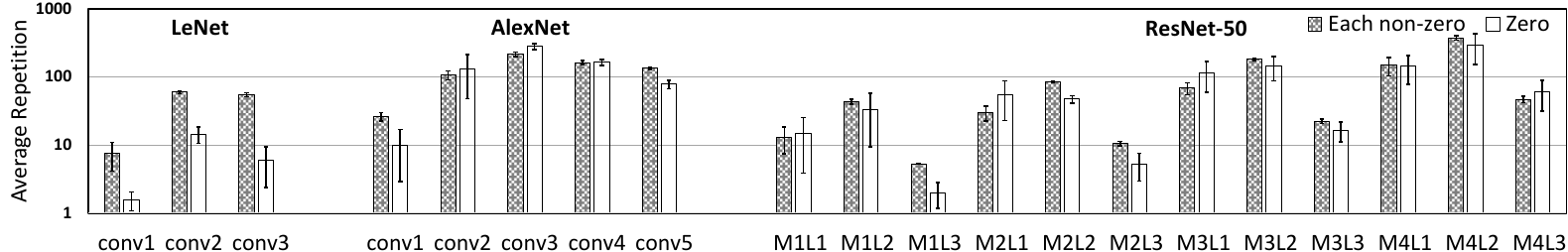}
  \caption{\label{fig:repetition_motivation}
            \small 
            Weight repetition per filter, averaged across all filters, for select layers in a Lenet-like CNN~\cite{caffecifar},  AlexNet~\cite{Alexnet} and  ResNet-50~\cite{Resnet}. All networks are trained with INQ~\cite{INQ}.
            LeNet was trained on CIFAR-10~\cite{cifar} and AlexNet/ResNet were trained on ImageNet~\cite{Imagenet}.
            MxLy stands for ``module x, layer y.'' 
            In the case of ResNet, we show one instance of each module, where repetition is averaged across filters in the layer. 
            Note that the error bars represent the standard deviation of weight repetition in each layer.
            } 
  \end{centering}
\end{figure*}


\subsection{CNN Background}

CNNs are made up of multiple layers, where the predominant layer type is a multi-dimensional convolution.
Each convolutional layer involves a 3-dimensional ($W\times H \times C$) input and $K$ 3-dimensional ($R\times S\times C$) filters.
Convolutions between the filters and input form a 3-dimensional $(W-R+1) \times (H-S+1) \times K$ output.
These parameters are visualized in Figure~\ref{fig:conv_params}.
$C$ and $K$ denote the layer's input and output channel count, respectively.
We will omit `$\times$' from dimensions for brevity when possible, e.g., $W\times H \times C\rightarrow WHC$.

\para{CNN inference.}
As with prior work~\cite{SCNN,Cambriconx,Cnvlutin}, this paper focuses on CNN inference, which is the online portion of the CNN computation run on, e.g., edge devices.
The inference operation for convolutional layers (not showing bias terms, and for unit stride) is given by
\begin{align}
&\out[(k, x, y)] = 
    \sum_{c=0}^{C-1} \sum_{r=0}^{R-1} \sum_{s=0}^{S-1} 
        \filt[(k, c, r, s)] * \inp[(c, x+r, y+s)] \label{eqn:conv} \\
    \nonumber &0 \le k < K, 0 \le x < W-R+1, 0 \le y < H-S+1
\end{align}
where $\out$, $\inp$ and $\filt$ are outputs (activations), inputs (activations) and filters (weights), respectively.
Outputs become inputs to the next layer.
Looking up $\out$, $\inp$ and $\filt$ with a tuple is notation for a multi-dimensional array lookup.
As is the case with other works targeting inference~\cite{EIE}, we assume a batch size of one.

We remark that CNNs have several other layer types including non-linear scaling~\cite{BatchNorm} layers, down-sampling/pooling layers and fully connected layers.
We focus on accelerating convolutional layers as they constitute the majority of the computation~\cite{Cong2014MinimizingCI}, but explain how to support other layer types in Section~\ref{sec:flexibility}.

\subsection{Weight Repetition in Modern CNNs}

We make a key observation that, while CNN filter dimensions have been relatively constant over time, the number of unique weights in each filter has decreased dramatically. 
This is largely due to several successful approaches to compress CNN model size~\cite{HanPTD15, precision, DeepCompression,INQ,Ternary}.
There have been two main trends, both referred to as \emph{weight quantization} schemes.
First, to decrease weight numerical precision, which reduces model size and the cost of arithmetic~\cite{TPU}. 
Second, to use a small set of high-precision weights~\cite{DeepCompression,INQ,Ternary}, which also reduces model size but can enable higher accuracy than simply reducing precision.

Many commercial CNNs today are trained with reduced, e.g., 8~bit~\cite{TPU}, precision per weight.
We refer to the number of unique weights in the network as $U$.
Thus, with 8~bit weights $U\le 2^8=256$.
Clearly, weight repetition within and across filters is guaranteed as long as $U < R*S*C$ and $U < R*S*C*K$, respectively.
This condition is common in contemporary CNNs, leading to a guaranteed weight repetition in modern networks. 
For example, every layer except the first layer in ResNet-50~\cite{Resnet} has more than 256 weights per filter and between $K=64$ to $K=512$ filters.

A complementary line of work shows it is possible to more dramatically reduce the number of unique weights, while maintaining state-of-the-art accuracy, by decoupling the number of unique weights from the numerical precision per weight~\cite{DeepCompression,INQ,Ternary}.
Figure~\ref{fig:repetition_motivation} shows weight repetition for several modern CNNs trained with a scheme called Incremental Network Quantization (INQ)~\cite{INQ}. 
INQ constrains the trained model to have only $U=17$ unique weights (16 non-zero weights plus zero) and achieves state-of-the-art accuracy on many contemporary CNNs.
Case in point, Figure~\ref{fig:repetition_motivation} shows a LeNet-like CNN from Caffe~\cite{caffecifar} trained on CIFAR-10~\cite{cifar}, and AlexNet~\cite{Alexnet} plus ResNet-50~\cite{Resnet} trained on ImageNet~\cite{Imagenet}, which achieved 80.16\%, 57.39\% and 74.81\% top-1 accuracy, respectively.

Figure~\ref{fig:repetition_motivation} shows that weight repetition is widespread and abundant across a range of networks of various sizes and depths.
We emphasize that repetition counts for the non-zero column in Figure~\ref{fig:repetition_motivation} are the average repetition for \emph{each} non-zero weight value \emph{within each} filter.
We see that each non-zero weight is seldom repeated less than 10 times.
Interestingly, the repetition count per non-zero is similar to that of the zero weight for most layers. 
This implies that the combined repetitions of non-zero weights (as there are $U-1$ non-zero weights) can dwarf the repetitions of zero weights.
The key takeaway message is that there is a large un-tapped potential opportunity to exploit repetitions in non-zero weights.

\section{Exploiting Weight Repetition}
\label{sec:comp_reuse}

We now discuss opportunities that leverage weight repetition to reduce work (save energy and cycles), based on refactoring and reusing CNN sub-computations.
From Section~\ref{sec:background}, there are $K$ CNN filters per layer, each of which spans the three dimensions of $RSC$.
Recall, $R$ and $S$ denote the filter's spatial dimensions and $C$ denotes the filter's channels.

We first present \emph{dot product factorization} (Section~\ref{sec:dpf}),
which saves multiplies by leveraging repeated weights within a single filter, i.e., throughout the $RSC$ dimensions.
We then present a generalization of dot product factorization, called \emph{activation group reuse} (Section~\ref{sec:agr}), to exploit repetition within and across filters, i.e., throughout the $RSCK$ dimensions.
Lastly, we remark on a third type of reuse that we do not exploit in this paper (Section~\ref{sec:ppr}) but may be of independent interest.

\begin{table}
{\small
\centering
  \caption{\label{tab:params}\footnotesize
    UCNN parameters. }
\begin{adjustwidth}{-.125cm}{}    
  \begin{tabular}{l | l | c}
    \hline 
    Name                     &   Description                & Defined \\
    \hline
    $U$          & Number of unique weights per CNN layer   & \ref{sec:background} \\
    $\indtable$ & Indirection table into input buffer      & \ref{sec:dpf} \\
    $\wttable$ & Indirection table into filter buffer     & \ref{sec:dpf} \\
    $G$          & Number of filters grouped for act. group reuse       & \ref{sec:agr} \\
    \hline 
  \end{tabular}
  \end{adjustwidth}
    }
\end{table}

For clarity, we have consolidated parameters and terminology unique to this paper in Table~\ref{tab:params}.

\subsection{Dot Product Factorization}
\label{sec:dpf}

Given each dot product in the CNN (an $RSC$-shape filter MACed to an $RSC$ sub-region of input), our goal is to reduce the number of multiplies needed to evaluate that dot product. 
This can be accomplished by factorizing out common weights in the dot product, as shown in Figure~\ref{fig:schemes}b.
That is, input activations that will be multiplied with the same weight
(e.g., $x, z$ and $y, k$ and $z, l$ in Figure~\ref{fig:schemes}b) are grouped and summed locally, and only that sum is multiplied to the repeated weight.
We refer to groups of activations summed locally as \emph{activation groups} --- we use this term extensively in the rest of the paper.
To summarize:
\begin{enumerate}
    \item Each activation group corresponds to one unique weight in the given filter.
    \item The total number of activation groups per filter is equal to the number of unique weights in that filter.
    \item The size of each activation group is equal to the repetition count for that group's weight in that filter.
\end{enumerate}
We can now express dot product factorization by rewriting the Equation~\ref{eqn:conv} as
\begin{align}
\out[(k, x, y)] = 
    \sum_{i=0}^{U} \left(\filt[\wttable[(k, i)]] * 
    \sum_{j=0}^{\mathrm{gsz}(k,i)-1} \inp[\indtable[(k, i, j)]]\right) \label{eqn:conv_factor}
\end{align}

$\out$, $\filt$ and $\inp$ are outputs, filters and inputs from Equation~\ref{eqn:conv}, $\mathrm{gsz}(k,i)$ indicates the size of the $i$-th activation group for the $k$-th filter and $U$ represents the number of unique weights in the network (or network layer).
Note that each filter can have a different number of unique weights due to an irregular weight distribution.
That is, some activation groups may be ``empty'' for a given filter.
For simplicity, we assume each filter has $U$ activation groups in this section, and handle the empty corner case in Section~\ref{sec:agr_arch}.

Activation groups are spread out irregularly within each $RSC$ sub-region of input. 
Thus, we need an indirection table to map the locations of each activation that corresponds to the same unique weight. 
We call this an \emph{input indirection table}, referred to as $\indtable$. 
The table $\indtable$ reads out activations from the input space in activation group-order. That is, $\indtable[(k,i,0)]\dots \indtable[(k,i,\mathrm{gsz}(k,i)-1)]$ represents the indices in the input space corresponding to activations in the $i$-th activation group for filter $k$.


Correspondingly, we also need to determine which unique weight should be multiplied to each activation group. We store this information in a separate \emph{weight indirection table}, referred to as $\wttable$. 
$\wttable[(k, i)]$ points to the unique weight that must be multiplied to the $i$-th activation group for filter $k$. 
We emphasize that, since the weights are static for a given model, both of the above tables can be pre-generated offline.

\para{Savings.}
The primary benefit from factorizing the dot product is reduced multiplications per dot product. 
Through the above scheme, the number of multiplies per filter reduces to the number of unique weights in the filter (e.g., 17 for INQ~\cite{INQ}), regardless of the size of the filter or activation group.
Referring back to Figure~\ref{fig:repetition_motivation}, average multiplication savings would be the height of each bar, and this ranges from $5\times$ to $373\times$. As discussed in Section~\ref{sec:background}, even out-of-the-box networks are guaranteed to see savings.

An important special case is the zero weight, or when $\filt[\wttable[(k, i)]] = 0$.
Then, the inner loop to sum the activation group and the associated multiplication is skipped.

\para{Costs.}
The multiply and sparsity savings come at the cost of storage overhead for the input and weight indirection tables, which na\"ively are the size of the original dense weights, as well as the energy costs to lookup inputs/weights through these tables. We introduce several techniques to reduce the size of each of these tables (Sections~\ref{sec:fdp_arch} to \ref{sec:agr_arch}). 
We also amortize the cost of looking up these tables through a novel vectorization scheme (Section~\ref{sec:spatial_vec}).

\subsection{Activation Group Reuse}
\label{sec:agr}

\begin{figure}[]
  \begin{centering}
  \includegraphics{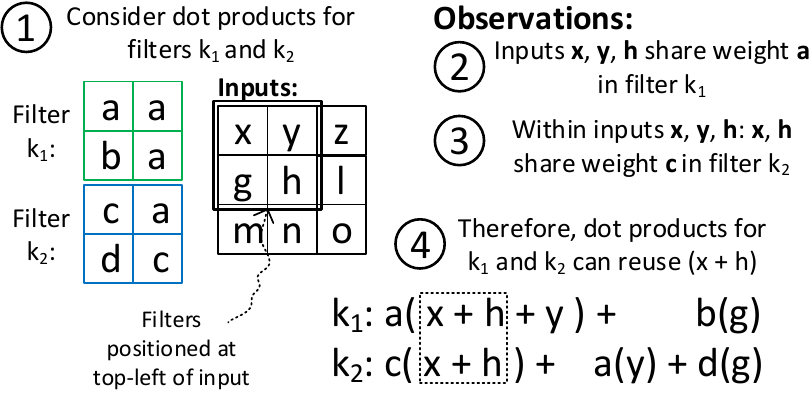}
  \caption{\label{fig:ag-reuse}
            \small 
            Activation group reuse example ($G=2$).
            }
  \end{centering}
\end{figure}

Dot product factorization reduces multiplications by exploiting weight repetition within the filter.
We can generalize the idea to simultaneously exploit repetitions across filters, using a scheme called \emph{activation group reuse}. 
The key idea is to exploit the overlap between two or more filters' activation groups. 
In Figure~\ref{fig:ag-reuse}, activations $x+h+y$ form an activation group for filter $k_1$.
Within this activation group, filter $k_2$ has a \emph{sub-activation group} which is $x+h$.  
The intersection of these two ($x+h$) can be reused across the two filters.

Formally, we can build the sub-activation groups for filter $k_2$, within filter $k_1$'s $i$-th activation group, as follows.
First, we build the activation group for $k_1$:
$$
A (k_1,i) = \{ \indtable[(k_1, i, j)]: j\in [0, \mathrm{gsz}(k_1, i)) \}
$$
Then, we build up to $U$ sub-activation groups for $k_2$ by taking set intersections.
That is, for $i'=0,\dots,U-1$, the $i'$-th sub-activation group for $k_2$ is given by:
$$
A(k_1, i) \bigcap A (k_2,i') 
$$

We can generalize the scheme to find overlaps across $G$ filters.
When $G=1$, we have vanilla dot product factorization (Section~\ref{sec:dpf}).
The discussion above is for $G=2$.
When $G>2$, we recursively form set intersections between filters $k_g$ and $k_{g+1}$, for $g=1,\dots,G-1$.
That is, once sub-activation groups for a filter $k_2$ are formed, we look for ``sub-sub'' activation groups within a filter $k_3$ which fall within the sub-groups for $k_2$, etc.
Formally, suppose we have a $g$th-level activation group $T_g$ for filter $k_g$. To find the $(g+1)$th-level activation groups for filter $k_{g+1}$ within $T_g$, we calculate $T_g \bigcap A(k_{g+1}, i')$ for $i'=0,\dots,U-1$, which is analogous to how intersections were formed for the $G=2$ case.

As mentioned previously, irregular weight distributions may mean that there are less than $U$ unique weights in filter $k_{g+1}$ that overlap with a given $g$th-level activation group for filter $k_g$.
We discuss how to handle this in Section~\ref{sec:agr_arch}.

\para{Savings.} 
Activation group reuse can bring significant improvements in two ways: 
\begin{enumerate}
    \item \textit{Reduced input buffer reads and arithmetic operations}: From Figure~\ref{fig:ag-reuse}, we can eliminate the buffer reads and additions for reused sub-expressions like $x+h$.
    The scheme simultaneously saves multiplies as done in vanilla dot product factorization.
    \item \textit{Compressed input indirection table $\indtable$}: Since we do not need to re-read the sub-, sub-sub-, etc. activation groups for filters $k_2,\dots,k_G$, we can reduce the size of the input indirection table $\indtable$ by an $O(G)$ factor. 
    We discuss this in detail in Section~\ref{sec:agr_arch}.
\end{enumerate}

\para{How prevalent is Activation Group Reuse?}
Activation group reuse is only possible when there are overlaps between the activation groups of two or more filters. 
If there are no overlaps, we cannot form compound sub-activation group expressions that can be reused across the filters. 
These overlaps are likely to occur when the filter size $R*S*C$ is larger than $U^G$, i.e., the number of unique weights to the $G$-th power. 
For example, for $(R,S,C) = (3,3,256)$ and $U = 8$, we expect to see overlaps between filter groups up to size $G=3$ filters.

We experimentally found that networks retrained with INQ~\cite{INQ} ($U=17$) and TTQ~\cite{Ternary} ($U=3$) can enable $G>1$. 
In particular, INQ satisfies between $G=2$ to 3 and TTQ satisfies $G=6$ to 7 for a majority of ResNet-50 layers. 
Note that these schemes can simultaneously achieve competitive classification accuracy relative to large $U$ schemes. 

\subsection{Partial Product Reuse}
\label{sec:ppr}

We make the following additional observation.
While dot product factorization looks for repetitions in each $RSC$-dimensional filter, it is also possible to exploit repetitions across filters, within the same input channel.
That is, across the $RSK$ dimensions for each input channel.
This idea is shown for 1D convolution in Figure~\ref{fig:schemes}c.
In CNNs, for each input channel $C$, if $w = \filt[(k_1, c, r_1, s_1)]=\filt[(k_2, c, r_2, s_2)]$ and $(k_1, r_1, s_1)\ne(k_2, r_2, s_2)$, partial products formed with weight $w$ can be reused across the filters, for the same spatial position, and as the filters slide.
We do not exploit this form of computation reuse further in this paper, as it is not directly compatible with the prior two techniques. 

\section{Processing Element Architecture}
\label{sec:architecture}

In this section, we will describe Processing Element (PE) architecture, which is the basic computational unit in the accelerator.
We will first describe the PE of an efficient Dense CNN accelerator, called DCNN.
We will then make PE-level changes to the DCNN design to exploit the weight repetition-based optimizations from Section~\ref{sec:comp_reuse}. 
This is intended to give a clear overview of how the UCNN design evolves over an efficient dense architecture and also to form a baseline for evaluations in  Section~\ref{sec:evaluation}.

\begin{figure}[h]
  \begin{centering}
  \includegraphics[width=.9\columnwidth]{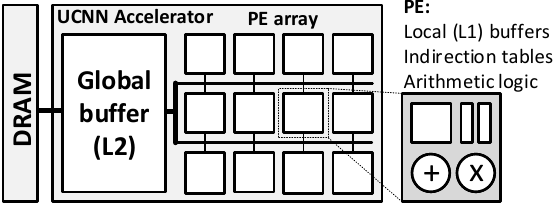}
  \caption{\label{fig:arch}
           \small 
           Chip-level DCNN/UCNN architecture.
           Indirection tables are UCNN only.
           } 
  \end{centering}
\end{figure}

The overall accelerator is made up of multiple PEs and a global buffer as depicted in Figure~\ref{fig:arch}.
The global buffer is responsible for scheduling work to the PEs.
We note that aside from changes to the PEs, the DCNN and UCNN accelerators (including their dataflow~\cite{Eyeriss}) are essentially the same.
We provide details on the overall (non-PE) architecture 
and dataflow in Section~\ref{sec:dataflow}. 


\subsection{Baseline Design: DCNN PE}
\label{sec:dcnn_pe}

The DCNN and UCNN PE's unit of work is to compute a dot product between an $RSC$ region of inputs and one or more filters. 
Recall that each dot product corresponds to all three loops in Equation~\ref{eqn:conv}, for a given $(k,x,y)$ tuple.

\begin{figure}[]
  \begin{centering}
  \includegraphics[width=\columnwidth]{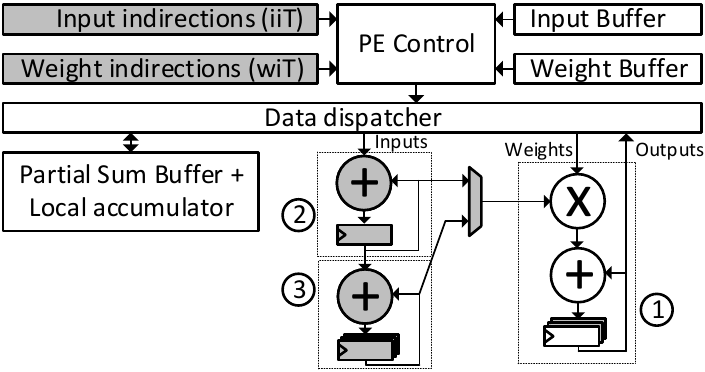}
  \caption{\label{fig:arch_all_scheme}
            \small 
            DCNN/UCNN PE Architecture. 
            Every component in grey is addition over the DCNN PE to design the UCNN PE. 
            \ding{192} represents a DCNN vector lane and \ding{192}-\ding{194} represents a UCNN vector lane. 
            \ding{193} is an accumulator added to sum activation groups for dot product factorization. 
            \ding{194} is an additional set of accumulators for storing sub-activation group partial sums.
            There are $G$ and $G-1$ accumulator registers in components \ding{192} and \ding{194}, respectively. } 
  \end{centering}
\end{figure}

To accomplish this task, the PE is made up of an input buffer, weight buffer, partial sum buffer, control logic and MAC unit (the non-grey components in  Figure~\ref{fig:arch_all_scheme}). 
At any point in time, the PE works on a filter region of size $RSC_t$ where $C_t\le C$, i.e., the filter is tiled in the channel dimension.
Once an $RSC_t$ region is processed, the PE will be given the next $RSC_t$ region until the whole $RSC$-sized dot product is complete.

Since this is a dense CNN PE, its operation is fairly straightforward. Every element of the filter is element-wise multiplied to every input element in the corresponding region, and the results are accumulated to provide a single partial sum. The partial sum is stored in the local partial sum buffer and is later accumulated with results of the dot products over the next $RSC_t$-size filter tile.

\para{Datapath.}
The datapath is made up of a fixed point multiplier and adder as shown in Figure~\ref{fig:arch_all_scheme} \ding{192}.
Once the data is available in the input and weight buffers, the control unit feeds the datapath with a weight and input element every cycle. They are MACed into a register that stores a partial sum over the convolution operation before writing back to the partial sum buffer. Together, we refer to this scalar datapath as a \emph{DCNN lane}.





\para{Vectorization.}
There are multiple strategies to vectorize this PE.
For example, we can vectorize across output channels (amortizing input buffer reads) by replicating the lane and growing the weight buffer capacity and output bus width.
DCNN and UCNN will favor different vectorization strategies, and we specify strategies for each later in the section and in the evaluation.

\subsection{Dot Product Factorization}
\label{sec:fdp_arch}

We now describe how to modify the DCNN architecture to exploit dot product factorization (Section~\ref{sec:dpf}).
The UCNN PE design retains the basic design and components of the DCNN PE along with its dataflow (Section~\ref{sec:dcnn_pe}). 
As described by Equation~\ref{eqn:conv_factor}, now the dot product operation is broken down into two separate steps in hardware:
\begin{enumerate}
    \item An inner loop which sums all activations within an activation group.
    \label{step:dpf:1}
    \item An outer loop which multiplies the sum from Step~\ref{step:dpf:1} with the associated weight and accumulates the result into the register storing the partial sum.
    \label{step:dpf:2}
\end{enumerate}

\para{Indirection table sorting.}
Compared to DCNN, we now additionally need two indirection tables: the input indirection table ($\indtable$) and the weight indirection table ($\wttable$) as discussed in Section~\ref{sec:dpf}.
Since we work on an $RSC_t$-size tile at a time, we need to load $RSC_t$ entries from both indirection tables into the PE at a time.
Following Equation~\ref{eqn:conv_factor} directly, each entry in $\indtable$ and $\wttable$ is a $\lceil \log_2 RSC_t \rceil$ and $\lceil \log_2 U\rceil$-bit pointer, respectively.

To reduce the size of these indirection tables and to simplify the datapath, we sort entries in the input and weight indirection tables such that reading the input indirections sequentially looks up the input buffer in activation group-order.
The weight indirection table is read in the same order. 
Note that because sorting is a function of weight repetitions, it can be performed offline.

Importantly, the sorted order implies that each weight in the weight buffer need only be read out once per activation group, and that the weight indirection table can be implemented as a single bit per entry (called the \emph{group transition} bit), to indicate the completion of an activation group.
Specifically, the next entry in the weight buffer is read whenever the group transition bit is set and the weight buffer need only store $U$ entries.


As mentioned in Section~\ref{sec:dpf}, we don't store indirection table entries that are associated with the zero weight. 
To skip zeros, we sort the zero weight to the last position and encode a ``filter done'' message in the existing table bits when we make the group transition to zero.
This lets UCNN skip zero weights as proposed by previous works~\cite{Cambriconx, SCNN} and makes the exploitation of weight sparsity a special case of weight repetition.



\para{Datapath.}
The pipeline follows the two steps from the beginning of the section, and requires another accumulator to store the activation group sum as reflected in Figure~\ref{fig:arch_all_scheme} \ding{193}.
As described above, the sorted input and weight indirection tables are read sequentially.
During each cycle in Step~\ref{step:dpf:1}, the input buffer is looked up based on the current input indirection table entry, and summed in accumulator \ding{193} until a group transition bit is encountered in the weight indirection table.
In Step~\ref{step:dpf:2}, the next weight from the weight buffer is multiplied to the sum in the MAC unit (Figure~\ref{fig:arch_all_scheme} \ding{192}). 
After every activation group, the pipeline performs a similar procedure using the next element from the weight buffer.

\para{Arithmetic bitwidth.}
This design performs additions before each multiply, which means the input operand in the multiplier will be wider than the weight operand. 
The worst case scenario happens when the activation group size is the entire input tile, i.e., the entire tile corresponds to one unique non-zero weight, in which case the input operand is widest.
This case is unlikely in practice, and increases multiplier cost in the common case where the activation group size is $\ll$ input tile size. 
Therefore, we set a maximum limit for the activation group size. In case the activation group size exceeds the limit, we split activation groups into chunks up to the maximum size.
A local counter triggers early MACs along with weight buffer `peeks' at group boundaries.
In this work, we assume a maximum activation group size of 16. 
This means we can reduce multiplies by $16\times$ in the best case, and the multiplier is 4~bits wider on one input.



\subsection{Activation Group Reuse}
\label{sec:agr_arch}

We now augment the above architecture to exploit activation group reuse (Section~\ref{sec:agr}).
The key idea is that by carefully ordering the entries in the input indirection table ($\indtable$), a single input indirection table can be shared across multiple filters.
This has two benefits.
First, we reduce the model size since the total storage needed for the input indirection tables shrinks.
Second, with careful engineering, we can share sub-computations between the filters, which saves input buffer reads and improves PE throughput. 
Recall that the number of filters sharing the same indirection table is a parameter $G$, as noted in Table~\ref{tab:params}.
If $G=1$, we have vanilla dot product factorization from the previous section.

\begin{figure}[h]
  \begin{centering}
  \includegraphics[width=\columnwidth]{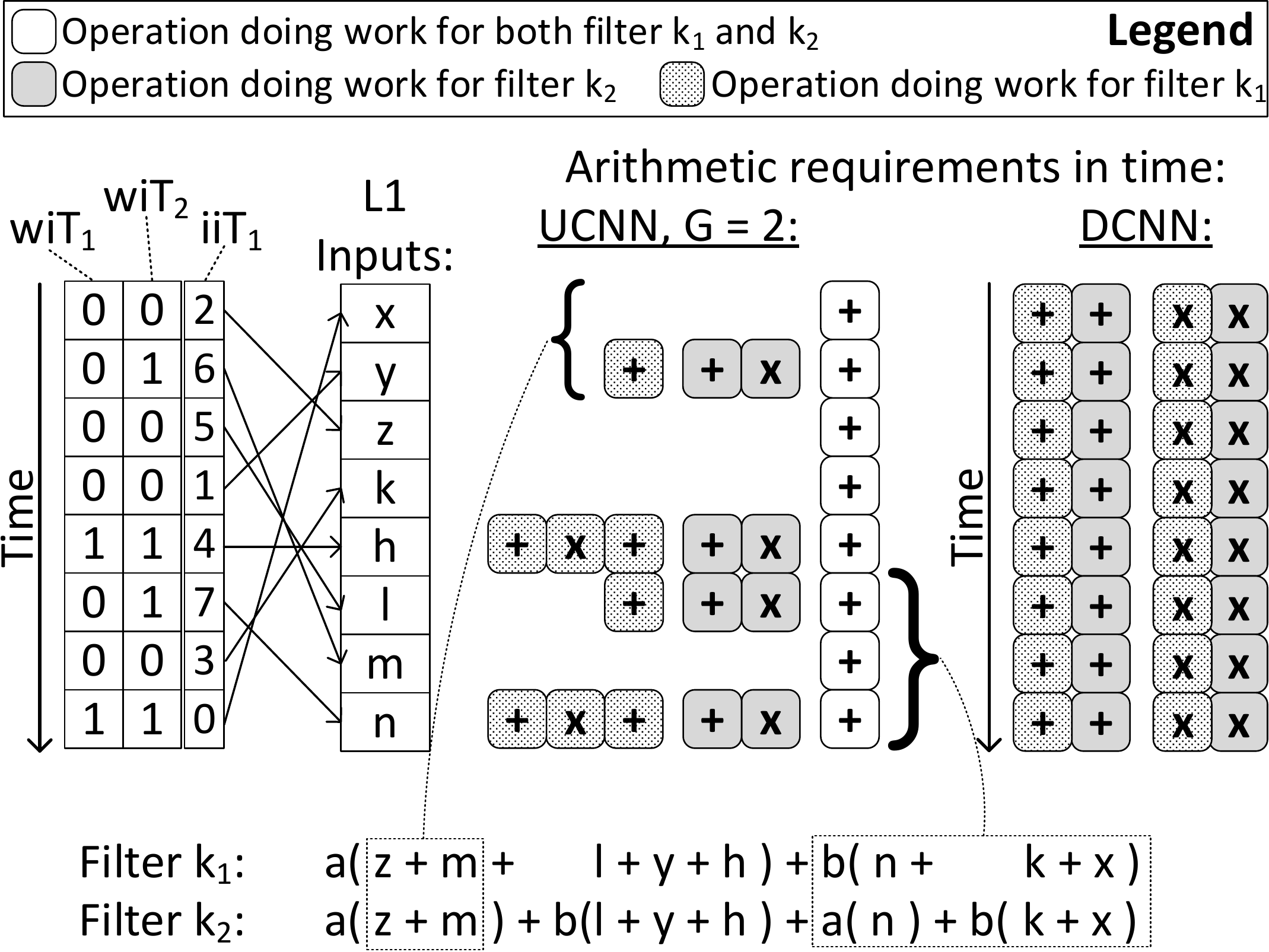}
  \caption{\label{fig:act_reuse_pipeline_example}
           \small 
           Example of activation group reuse for $G=2$ with weights $a,b$.
           The indirection tables $\indtable$ and $\wttable$ are walked top to bottom (time moves down).
           At each step, the sequence of adds and multiplies needed to evaluate that step are shown right to left.
           Recall a MAC is a multiply followed by an add.
           We assume that at the start of building each sub-activation group, the state for accumulator~\ding{193} is reset to 0.
           As shown, DCNN with two DCNN lanes processes these filters with 16 multiplies, whereas UCNN completes the same work in 6 multiplies.
           } 
  \end{centering}
\end{figure}

\para{Indirection table hierarchical sorting.}
To support $G>1$, we hierarchically sort a single input indirection table to support $G$ filters.
Due to the hierarchical sort, we will still be able to implement the weight indirection tables as a single bit per entry per filter, as done in Section~\ref{sec:fdp_arch}.
We give an example for the $G=2$ case in Figure~\ref{fig:act_reuse_pipeline_example}, and walk through how to hierarchically sort the indirection tables below.
These steps are performed offline.
\begin{enumerate}
    \item   Select a canonical order of weights.
            The order is $a,b$ in the example.
            \label{step1:S43}
    \item   Sort entries by activation group for the first filter $k_1$.
            \label{step2:S43}
    \item   Within each activation group of $k_1$, sort by sub-activation group for the second filter $k_2$ using the \emph{same} canonical order $a,b$. 
            Filter $k_1$ has activation groups (e.g., $z + m + l + y + h$) and filter $k_2$ has sub-activation groups within filter $k_1$'s groups (e.g., $z + m $ and $l+y+h$). 
            \label{step3:S43}
    
\end{enumerate}
 
Now, a single traversal of the input indirection table can efficiently produce results for both filters $k_1$ and $k_2$.
Crucially, sorts performed in Step~\ref{step2:S43} and \ref{step3:S43} are \emph{keyed to the same canonical order of weights} we chose in Step~\ref{step1:S43} ($a,b$ in the example).
By keeping the same order across filters, the weight indirection tables (denoted $\wttable_1$ and $\wttable_2$ for $k_1$ and $k_2$, respectively, in Figure~\ref{fig:act_reuse_pipeline_example}) can be implemented as a single bit per entry. 

As mentioned above, the scheme generalizes to $G>2$ in the natural fashion.
For example, for $G=3$ we additionally sort sub-sub-activation groups within the already established sub-activation groups using the same canonical $a,b$ weight order.
Thus, the effective indirection table size per weight is $(|\indtable.\mathsf{entry}| + G * |\wttable.\mathsf{entry}|)/G = \lceil \log_2 RSC_t \rceil/G + 1$ which is an $O(G)$ factor compression.
We will see the upper bound for $G$ later in this section.

\para{Datapath.}
To support activation group reuse, we add a third accumulator to the PE to enable accumulations across different level activation groups (Figure~\ref{fig:arch_all_scheme} \ding{194}). 
$G$-th activation groups are first summed in accumulator~\ding{193}.
At $G$-th level activation group boundaries, the $G$-th level sum is merged into running sums for levels $g=1,\dots,G-1$ using accumulator~\ding{194}. 
At any level activation group boundary,  sums requiring a multiply are dispatched to the MAC unit~\ding{192}.


For clarity, we now give a step-by-step (in time) description of this scheme using the example in Figure~\ref{fig:act_reuse_pipeline_example} and the architecture from Figure~\ref{fig:arch_all_scheme}.
Recall, we will form activation groups for filter $k_1$ and sub-activation groups for filter $k_2$.
\begin{enumerate}
    \item The input indirection table $\indtable$ reads the indirection to be $2$, which corresponds to activation $z$. 
          This is sent to accumulator~\ding{193} which starts building the sub-activation group containing $z$.
          We assume accumulator~\ding{193}'s state is reset at the start of each sub-activation group, so the accumulator implicitly calculates $0+z$ here.
          Both $\wttable_1$ and $\wttable_2$ read $0$s, thus we proceed without further accumulations.
    \item $\indtable$ reads $6$ and $\wttable_1$ and $\wttable_2$ read $0$ and $1$, respectively. 
          This means we are at the end of the sub-activation group (for filter $k_2$), but not the activation group (for filter $k_1$). 
          Sum $z+m$ is formed in accumulator~\ding{193}, which is sent (1) to accumulator~\ding{194}---as this represents the sum of only a part of the activation group for filter $k_1$---and (2) to the MAC unit~\ding{192} to multiply with $a$ for filter $k_2$.
    \item Both $\wttable_1$ and $\wttable_2$ read $0$s, accumulator~\ding{193} starts accumulating the sub-activation group containing $l$.
    \item Both $\wttable_1$ and $\wttable_2$ read $0$s, accumulator~\ding{193} builds $l+y$.
    \item Both $\wttable_1$ and $\wttable_2$ read $1$s, signifying the end of both the sub-activation and activation groups. 
          Accumulator~\ding{193} calculates $l+y+h$, while accumulator~\ding{194} contains $z+m$ for filter $k_1$. 
          The result from accumulator~\ding{193} is sent (1) to the MAC Unit~\ding{192}---to multiply with $b$ for filter $k_2$---and (2) to accumulator~\ding{194} to generate $z+m+l+y+h$. 
          The result from accumulator~\ding{194} finally reaches the MAC Unit~\ding{192} to be multiplied with $a$.
    \item Repeat steps similar to those shown above for subsequent activation groups on filter $k_1$, until the end of the input indirection table traversal.
\end{enumerate}

Together, we refer to all of the above arithmetic and control as a \emph{UCNN lane}. Note that a transition between activation groups in $k_1$ implies a transition for $k_2$ as well.

\para{Area implications.}
To vectorize by a factor of $G$, a dense design requires $G$ multipliers. 
However, as shown in Figure~\ref{fig:arch_all_scheme}, we manage to achieve similar throughput with a single multiplier.
The multiplier reduction is possible because the multiplier is only used on (sub-)activation group transitions.
We do note that under-provisioning multipliers can lead to stalls, e.g., if (sub-)activation group transitions are very common.
Thus, how many hardware multipliers and accumulators to provision is a design parameter. 
We evaluate a single-multiplier design in Section~\ref{sec:jiyong}. 


\para{Handling empty sub-activation groups.}
In Figure~\ref{fig:act_reuse_pipeline_example}, if weight $a$ or $b$ in filters $k_1$ or $k_2$ had a (sub-)activation group size of zero, the scheme breaks because each filter cycles through weights in the same canonical order. 
To properly handle these cases, we have two options.
First, we can allocate more bits per entry in the weight indirection table.
That is, interpret weight indirection table entries as $n$-bit counters that can skip 0 to $2^n-1$ weights per entry.
Second, we can add special ``skip'' entries to the weight and input indirection tables to skip the weight without any computations. 
A simple skip-entry design would create a cycle bubble in the UCNN lane per skip.

We apply a hybrid of the above schemes in our implementation.
We provision an extra bit to each entry in the $G$-th filter's weight indirection table, for each group of $G$ filters.
An extra bit enables us to skip up to 3 weights.
We find we only need to add a bit to the $G$-th filter, as this filter will have the smallest activation groups and hence has the largest chance of seeing an empty group.
For any skip distance longer than what can be handled in allocated bits, we add skip entries as necessary and incur pipeline bubbles. 

\para{Additional table compression.}
We can further reduce the bits per entry in the input indirection table by treating each entry as a jump, relative to the last activation sharing the same weight, instead of as a direct pointer.
This is similar to run-length encodings (RLEs) in sparse architectures~\cite{Eyeriss,EIE,SCNN}.
Represented as jumps, bits per table entry are proportional to the average distance between activations sharing the same weight (i.e., $O(\log_2 U)$), which can be smaller than the original pointer width $\lceil \log_2 RSC_t \rceil$.
The trade-off with this scheme is that if the required jump is larger than the bits provisioned, we must add skip entries to close the distance in multiple hops.\footnote{Similar issues are faced by RLEs for sparsity~\cite{EIE,Eyeriss}.}

\para{Activation group reuse implications for weight sparsity.}
Fundamentally, to service $G$ filters we need to read activations according to the \emph{union} of non-zero weights in the group of $G$ filters.
That is, we can only remove entries from indirection tables if the corresponding weight in filters $k_1$ \emph{and} $k_2$ is 0.
Thus, while we get an $O(G)$ factor of compression in indirection tables, less entries will be skip-able due to weight sparsity.

\subsection{Spatial Vectorization}
\label{sec:spatial_vec}

One overhead unique to the UCNN PE is the cost to indirect into the input buffer.
The indirection requires an extra buffer access, and the irregular access pattern means the input SRAM cannot read out vectors (which increases pJ/bit).
Based on the observation that indirection tables are reused for every filter slide, we propose a novel method to vectorize the UCNN PE across the spatial $WH$ dimensions.
Such reuse allows UCNN to amortize the indirection table lookups across vector lanes. 
We refer to this scheme as \emph{spatial vectorization} and introduce a new parameter $V_W$ to indicate the spatial vector size.

To implement spatial vectorization, we split the input buffer into $V_W$ banks and carefully architect the buffer so that exactly $V_W$ activations can be read every cycle. 
We note the total input buffer capacity required is only $O(C_t * S * (V_W+R))$, not $O(C_t * S * V_W * R)$, owing to the overlap of successive filter slides.
The datapath for activation group reuse (Section~\ref{sec:agr_arch}) is replicated across vector lanes, thus improving the  PE throughput to $O(G*V_W)$ relative to the baseline non-vectorized PE. 
Given that UCNN significantly reduces multiplier utilization, an aggressive implementation could choose to temporally multiplex $<V_W$ multipliers instead of spatially replicating multipliers across lanes.


\para{Avoiding bank conflicts.}
Since the input buffer access pattern is irregular in UCNN, there may be bank conflicts in the banked input buffer. 
To avoid bank conflicts, we divide the input buffer into $V_W$ banks and apply the following fill/access strategy.  
To evaluate $V_W$ dot products, we iterate through the input buffer according to the input indirection table.
We denote each indirection as a tuple $(r,s,c)\in RSC_t$, where $(r,s,c)$ corresponds to the spatial vector base address. 
Then, the bank id/bank address to populate vector slot $v\in[0,\dots,V_W-1]$ for that indirection is:
\begin{align}
\mathrm{bank}(r,s,c,v)   &= (r + v) \; \% \; V_W \label{eqn:bankid} \\
\mathrm{addr}(r,s,c,v)   &= s * C_t + c + \big\lceil(r + v) / V_W\big\rceil * S * C_t  \label{eqn:bankaddr}
\end{align}
This strategy is bank conflict free because $\mathrm{bank(r,s,c,v)}$ always yields a different output for fixed $(r,s,c)$, varying $v$. 
Unfortunately, this scheme has a small storage overhead: a $((R+V_W-1) \; \% \; V_W) / (R+V_W-1)$ fraction of addresses in the input buffer are un-addressable.
Note, this space overhead is always $<2\times$ and specific settings of $R$ and $V_W$ can completely eliminate overhead (e.g., $V_W=2$ for $R=3$).



%


%
 
\subsection{UCNN Design Flexibility}
\label{sec:flexibility}

\para{Supporting a range of $U$.}
Based on the training procedure, CNNs may have a different number of unique weights (e.g., 3~\cite{Ternary} or 17~\cite{INQ} or 256~\cite{DeepCompression} or more).
Our accelerator can flexibly handle a large range of $U$, but still gain the efficiency in Section~\ref{sec:dcnn_pe}, by reserving a larger weight buffer in the PE. 
This enables UCNN to be used on networks that are not re-trained for quantization as well. 
We note that even if $U$ is large, we still save energy by removing redundant weight buffer accesses.


\para{Support for other layer types.}
CNNs are made up of multiple layer types including convolutional, non-linear activation, pooling and fully connected.
We perform non-linear activations (e.g., ReLu~\cite{Relu}) at the PE (see Figure~\ref{fig:dataflow} \texttt{(F)}).
Pooling can be handled with minimal additional logic (e.g., max circuits) at the PE, with arithmetic disabled.
We implement fully connected layers as convolutions where input buffer slide reuse is disabled (see next section). 

\section{Architecture and Dataflow}
\label{sec:dataflow}

This section presents the overall
architecture for DCNN and UCNN, i.e., components beyond the PEs, as well as the architecture's dataflow.
CNN dataflow~\cite{Eyeriss} describes how and when data moves through the chip.
We present a dataflow that both suits the requirements of UCNN and provides the best power efficiency and performance out of candidates 
that we tried. 

As described in the previous section and in Figure~\ref{fig:arch}, the DCNN and UCNN architectures consist of multiple Processing Elements (PEs) connected to a shared global buffer (L2), similar to previous proposals~\cite{DaDianNao,Eyeriss,Cambriconx}. 
Similar to the PEs, the L2 buffer is divided into input and weight buffers. 
When it is not clear from context, we will refer to the PE-level input and weight buffers (Section~\ref{sec:architecture}) as the L1 input and weight buffers.
Each PE is fed by two multicast buses, for input and weight data. 
Final output activations, generated by PEs, are written back to the L2 alongside the input activations in a double-buffered fashion.
That is, each output and will be treated as an input to the next layer. 

\subsection{Dataflow}

Our dataflow is summarized as follows.
We adopt weight- and output-stationary terminology from \cite{Eyeriss}.
\begin{enumerate}
    \item The design is weight-stationary at the L2, and stores all input activations on chip when possible.
    \item Each PE produces one column of output activations and PEs work on adjacent overlapped regions of input.
    The overlapping columns create input halos~\cite{SCNN}.
    \item Each PE is output-stationary, i.e., the partial sum resides in the PE until the final output is generated across all $C$ input channels.
\end{enumerate}

At the top level, our dataflow strives to minimize reads/writes to DRAM as DRAM often is the energy bottleneck in CNN accelerators~\cite{Cambriconx,SCNN}.
Whenever possible, we store all input activations in the L2.
We do not write/read input activations from DRAM unless their size is prohibitive.
We note that inputs fit on chip in most cases, given several hundred KB of L2 storage.\footnote{For example, all but several ResNet-50~\cite{Resnet} layers can fit inputs on chip with 256~KB of storage and 8~bit activations.}
In cases where inputs fit, we only need to read inputs from DRAM once, during the first layer of inference. 
In cases where inputs do not fit, we tile the input spatially.
In all cases, we read all weights from DRAM for every layer.
This is fundamental given the large (sometimes 10s of MB) aggregate model size counting all layers. 
To minimize DRAM energy from weights, the dataflow ensures that each weight value is fetched a minimal number of times, e.g., once if inputs fit and once per input tile otherwise. 

At the PE, our dataflow was influenced by the requirements of UCNN. 
Dot product factorization (Section~\ref{sec:comp_reuse}) builds activation groups through $RSC$ regions, hence the dataflow is designed to give PEs visibility to $RSC$ regions of weights and inputs in the inner-most (PE-level) loops. 
We remark that dataflows working over $RSC$ regions in the PEs have other benefits, such as reduced partial sum movement~\cite{Eyeriss,DaDianNao}.

Detailed pseudo-code for the complete dataflow is given in Figure~\ref{fig:dataflow}.
For simplicity, we assume the PE is not vectorized.
Inputs reside on-chip, but weights are progressively fetched from DRAM in chunks of $K_c$ filters at a time \texttt{(A)}. 
$K_c$ may change from layer to layer and is chosen such that the L2 is filled.
Work is assigned to the PEs across columns of input and filters within the $K_c$-size group \texttt{(B)}.
Columns of input and filters are streamed to PE-local L1 buffers \texttt{(C)}.
Both inputs and weights may be multicast to PEs (as shown by \texttt{\#multicast}), depending on DNN layer parameters.
As discussed in Section~\ref{sec:dcnn_pe}, $C_t$ input channels-worth of inputs and weights are loaded into the PE at a time. 
As soon as the required inputs/weights are available, $RSC_t$ sub-regions of input are transferred to smaller L0 buffers for spatial/slide data reuse and the dot product is calculated for the $RSC_t$-size tile \texttt{(E)}. 
Note that each PE works on a column of input of size $RHC$ and produces a column of output of size $H$ \texttt{(D)}. 
The partial sum produced is stored in the L1 partial sum buffer and the final output is written back to the L2 \texttt{(F)}. 
Note that the partial sum resides in the PE until the final output is generated, making the PE dataflow output-stationary.

\begin{figure}[ht!]
  \begin{centering}

  \begin{lstlisting}
    def CNNLayer():
      BUFFER in_L2 [C][H][W]; 
      BUFFER out_L2[K][H][W]; 
      BUFFER wt_L2 [Kc][C][S][R];
(A)   for kc = 0 to K/Kc - 1 
      {              
        wt_L2 = DRAM[kc*Kc:(kc+1)Kc-1]
                    [:][:][:];
        #parallel
(B)     for col, k in (col = 0 to W-R) x 
                      (k = 0 to Kc-1) 
        {
          PE(col, k);
        }
      }
      
    def PE(col, k):
      // col: which spatial column
      // k: filter
      BUFFER in_L1  [Ct][S][R];
      BUFFER psum_L1[H];
      BUFFER wt_L1  [Ct][S][R];
      psum_L1.zero(); // reset psums
(C)   for ct = 0 to C/Ct - 1
      {
        #multicast
        wt_L1 = wt_L2[k][ct*Ct:(ct+1)Ct-1]
                     [:][:];
(D)     for h = 0 to H - S
        {
          // slide reuse for in_L1 not shown
          #multicast
          in_L1 = in_L2[ct*Ct:(ct+1)Ct-1]
                       [h:h+S-1]
                       [col:col+R-1];
          sum = psum_L1[h];
(E)       for r,c,s in (r = 0 to R-1) x     
                       (c = 0 to Ct-1) x 
                       (s = 0 to S-1)
          {
            act = in_L1[c][s][r];
            wt = wt_L1[c][s][r];
            sum += act * wt;
          }
          psum_L1[h] = sum;
        }
      }
(F)   out_L2[k][:][col] = RELU(psum_L1);                          
  \end{lstlisting}

  \caption{\label{fig:dataflow}
           \small 
           DCNN/UCNN dataflow, parameterized for DCNN~(Section~\ref{sec:dcnn_pe}).
           For simplicity, the PE is not vectorized and stride is assumed to be 1.
           \texttt{[x:y]} indicates a range; \texttt{[:]} implies all data in that dimension.
           }
  \end{centering}
\end{figure}

\section{Evaluation}
\label{sec:evaluation}


\subsection{Methodology}
\label{sec:methods}

\para{Measurement setup.}
We evaluate UCNN using a whole-chip performance and energy model, and design/synthesize the DCNN/UCNN PEs in RTL written in Verilog.
All designs are evaluated in a 32~nm process, assuming a 1~GHz clock.
For the energy model, energy numbers for arithmetic units are taken from~\cite{marktalk}, scaled to 32~nm.
SRAM energies are taken from CACTI~\cite{cacti}.
For all SRAMs, we assume \texttt{itrs-lop} as this decreases energy per access, but still yields SRAMs that meet timing at 1~GHz.
DRAM energy is counted at 20~pJ/bit~\cite{marktalk}.
Network on chip (NoC) energy is extrapolated based on the number and estimated length of wires in the design (using our PE area and L2 SRAM area estimates from CACTI).
We assume the NoC uses low-swing wires~\cite{cacti-lowswing}, which are low power, however consume energy each cycle (regardless of whether data is transferred) via differential signaling.

\para{Activation/weight data types.}
Current literature employs a variety of activation/weight precision settings.
For example, 8 to 16~bit fixed point~\cite{SCNN,EIE,DaDianNao,TPU,DeepCompression}, 32~bit floating point/4~bit fixed point activations with power of two weights~\cite{INQ} to an un-specified (presumably 16~bit fixed point) precision~\cite{Ternary}. 
Exploiting weight repetition is orthogonal to which precision/data type is used for weights and activations.
However, for completeness, we evaluate both 8~bit and 16~bit fixed point configurations.

\para{Points of comparison.}
We evaluate the following design variants: 

\begin{enumerate}
\item [$\mathsf{DCNN}$:]
Baseline DCNN (Section~\ref{sec:dcnn_pe}) that does not exploit weight or activation sparsity, or weight repetition.
We assume that DCNN is vectorized across output channels and denote the vector width as $V_k$.
Such a design amortizes the L1 input buffer cost and improves DCNN's throughput by a factor of $V_k$.

\item [$\mathsf{DCNN\_sp}$:]
DCNN with Eyeriss-style~\cite{Eyeriss} sparsity optimizations.
\conf{DCNN\_sp} skips multiplies at the PEs when an operand (weight or activation) is zero, and compresses data stored in DRAM with a 5~bit run-length encoding.


\item [$\mathsf{UCNN\_Uxx}$:]
UCNN, with all optimizations enabled (Section~\ref{sec:agr_arch}) except for the jump-style indirection table (Section~\ref{sec:agr_arch}) which we evaluate separately in Section~\ref{sec:model_size}.
UCNN reduces DRAM accesses based on weight sparsity and activation group reuse, and reduces input memory reads, weight memory reads, multiplies and adds per dot product at the PEs.
UCNN also vectorizes spatially (Section~\ref{sec:spatial_vec}).
The \conf{\_Uxx} refers to the number of unique weights; e.g., \conf{UCNN\_U17} is UCNN with $U=17$ unique weights, which corresponds to an INQ-like quantization.
\end{enumerate}


\para{CNNs evaluated.}
To prove the effectiveness of UCNN across a range of contemporary CNNs, we evaluate the above schemes on three popular CNNs: a LeNet-like CNN~\cite{caffecifar} trained on CIFAR-10~\cite{caffecifar}, and AlexNet~\cite{Alexnet} plus ResNet-50~\cite{Resnet} trained on ImageNet~\cite{Imagenet}.
We refer to these three networks as LeNet, AlexNet and ResNet for short.


\subsection{Energy Analysis}
\label{sec:eval_fluid}


We now perform a detailed energy analysis and design space exploration comparing DCNN and UCNN.  

\para{Design space.}
We present results for several weight density points (the fraction of weights that are non-zero), specifically 90\%, 65\% and 50\%.
For each density, we set (100-density)\% of weights to 0 and set the remaining weights to non-zero values via a uniform distribution.
Evaluation on a real weight distribution from INQ training is given in Section~\ref{sec:jiyong}.
90\% density closely approximates our INQ data.
65\% and 50\% density approximates prior work, which reports negligible accuracy loss for this degree of sparsification~\cite{DeepCompression,Ternary,SCNN}.
We note that UCNN does not alter weight values, hence UCNN run on prior training schemes~\cite{DeepCompression,INQ,Ternary} results in the same accuracy as reported in those works. 
Input activation density is 35\% (a rough average from \cite{SCNN}) for all experiments.
We note that lower input activation density favors \conf{DCNN\_sp} due to its multiplication skipping logic.

To illustrate a range of deployment scenarios, we evaluate UCNN for different values of unique weights: $U=3,17,64,256$. 
We evaluate \conf{UCNN\_U3} (``TTQ-like''~\cite{Ternary}) and \conf{UCNN\_U17} (``INQ-like''~\cite{INQ}) as these represent two example state-of-the-art quantization techniques. 
We show larger $U$ configurations to simulate a range of other quantization options.
For example, \conf{UCNN\_U256} can be used on out-of-the-box (not re-trained) networks quantized for 8~bit weights~\cite{TPU} or on networks output by Deep Compression with 16~bit weights~\cite{DeepCompression}.

\para{Hardware parameters.}
Table~\ref{tab:bc_hardware_params} lists all the  hardware parameters used by different schemes in this evaluation.
To get an apples-to-apples performance comparison, we equalize ``effective throughput'' across the designs in two steps.
First, we give each design the same number of PEs.
Second, we vectorize each design to perform the work of 8 dense multiplies per cycle per PE.
Specifically, DCNN uses $V_K=8$ and UCNN uses $V_W$ and $G$ such that $G*V_W=8$, where $V_W$ and $V_K$ represent vectorization in the spatial and output channel dimensions, respectively.
Note that to achieve this throughput, the UCNN PE may only require $V_W$ or fewer multipliers (Section~\ref{sec:agr_arch}).
Subject to these constraints, we allow each design variant to choose a different L1 input buffer size, $V_W$ and $G$ to maximize its own average energy efficiency.


\begin{table}
{\small
\begin{center}
  \caption{\label{tab:bc_hardware_params}\small
    UCNN, DCNN hardware parameters with memory sizes shown in bytes.
    For UCNN:
    L1 wt. (weight) is given as the sum of weight table storage $|\indtable|+|\wttable|+|\filt|$ (Section~\ref{sec:dpf}).
    }
  \begin{tabular}{l | c | c | c | c | c | c }
    \hline 
    Design                   &   $P$ & $V_K$ & $V_W$ & $G$ & L1 inp. & L1 wt. \\
    \hline
    \conf{DCNN}              &   32  & 8     & 1     & 1  & 144     & 1152 \\
    \conf{DCNN\_sp}          &   32  & 8     & 1     & 1  & 144     & 1152 \\
    \conf{UCNN} ($U=3$)      &   32  & 1     & 2     & 4  & 768     & 129 \\ 
    \conf{UCNN} ($U=17$)   &   32  & 1     & 4     & 2  & 1152    & 232 \\ 
    \conf{UCNN} ($U>17$)     &   32  & 1     & 8     & 1  & 1920    & 652 \\ 
    \hline 
  \end{tabular}
  \end{center}
}
  
\end{table} 

\begin{figure*}[h]
  \begin{centering}
  \includegraphics[width=\textwidth]{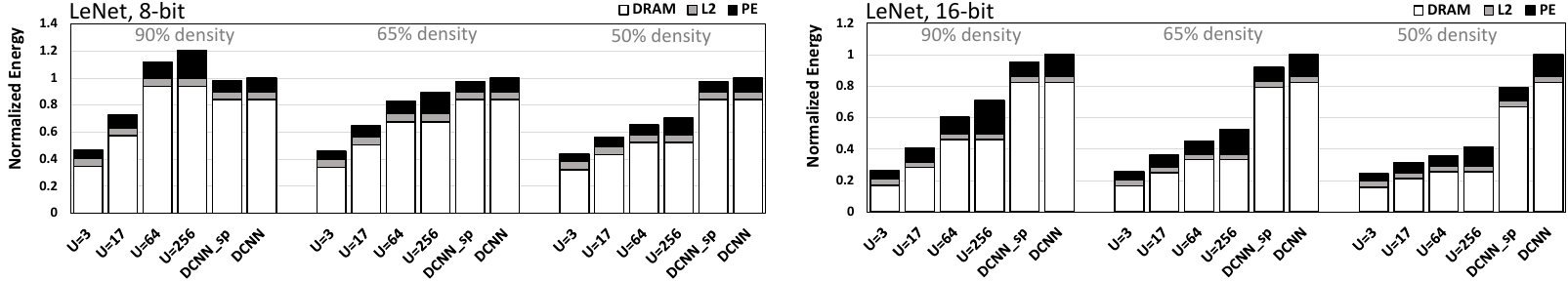}
  \includegraphics[width=\textwidth]{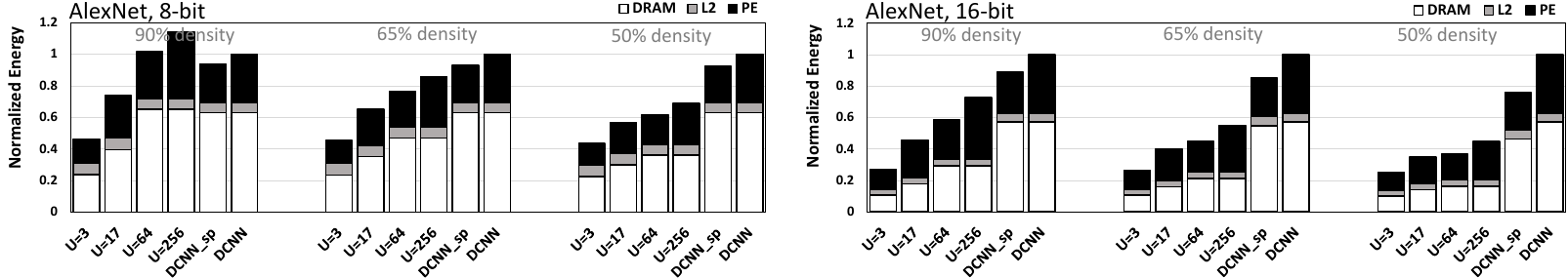}
  \includegraphics[width=\textwidth]{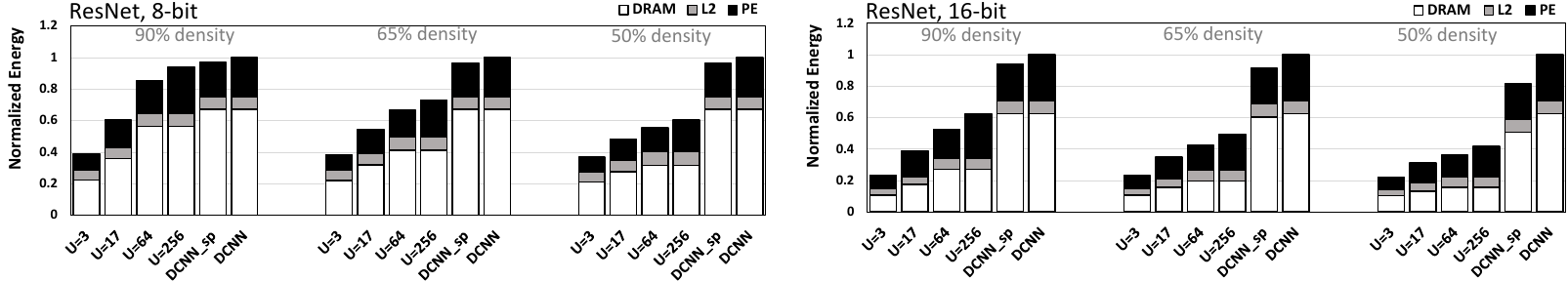}
  \caption{\label{eval:design_space_energy}
          \small 
          Energy consumption analysis of the three popular CNNs discussed in Section~\ref{sec:methods}, running on UCNN and DCNN variants.
          UCNN variant \conf{UCNN\_Uxx} is shown as $U=xx$.
          Left and right graphs show results using 8~bit and 16~bit weights, respectively.
          For each configuration, we look at 90\%, 65\% and 50\% weight densities.
          In all cases, input density is 35\%.
          Each group of results (for a given network and weight precision/density) is normalized to the \conf{DCNN} configuration in that group.
          } 
  \end{centering}
\end{figure*}


\para{Results.}
Figure~\ref{eval:design_space_energy} shows energy consumption for three contemporary CNNs at both 8 and 16 bit precision. 
Energy is broken into DRAM, L2/NoC and PE components.
Each configuration (for a particular network, weight precision and weight density) is normalized to \conf{DCNN} for that configuration.


At 16~bit precision, all UCNN variants reduce energy compared to \conf{DCNN\_sp}.
The improvement comes from three sources.
First, activation group reuse ($G>1$ designs in Table~\ref{tab:bc_hardware_params}) reduces DRAM energy by sharing input indirection tables across filters.
Second, activation group reuse (for any $G$) reduces energy from arithmetic logic at the PE.
Third, decreasing weight density results in fewer entries per indirection table on average, which saves DRAM accesses and cycles to evaluate each filter.
Combining these effects, \conf{UCNN\_U3}, \conf{UCNN\_U17} and \conf{UCNN\_U256} reduce energy by up to $3.7\times$, $2.6\times$ and $1.9\times$, respectively, relative to \conf{DCNN\_sp} for ResNet-50 at 50\% weight density.
We note that 50\% weight density improves \conf{DCNN\_sp}'s efficiency since it can also exploit sparsity.
Since \conf{DCNN} cannot exploit sparsity, UCNN's improvement widens to $4.5\times$, $3.2\times$ and $2.4\times$ compared to \conf{DCNN}, for the same configurations.
Interestingly, when given relatively dense weights (i.e., 90\% density as with INQ training), the UCNN configurations attain a $4\times$, $2.4\times$ and $1.5\times$ improvement over \conf{DCNN\_sp}.
The improvement for \conf{UCNN\_U3} increases relative to the 50\% dense case because \conf{DCNN\_sp} is less effective in the dense-weights regime.

We observed similar improvements for the other networks (AlexNet and LeNet) given 16~bit precision, and improvement across all networks ranges between $1.2\times\sim4\times$
and 
$1.7\times\sim3.7\times$
for 90\% and 50\% weight densities, respectively.

At 8~bit precision, multiplies are relatively cheap and DRAM compression is less effective due to the relative size of compression metadata.
Thus, improvement for \conf{UCNN\_U3}, \conf{UCNN\_U17} and \conf{UCNN\_U256} drops to $2.6\times$, $2\times$ and $1.6\times$, respectively, relative to \conf{DCNN\_sp} on ResNet-50 and 50\% weight density. 
At the $90\%$ weight density point, UCNN variants with $U=64$ and $U=256$ perform worse than \conf{DCNN\_sp} on AlexNet and LeNet. 
These schemes use $G=1$ and thus incur large energy overheads from reading indirection tables from memory.
We evaluate additional compression techniques to improve these configurations in Section~\ref{sec:model_size}.

\begin{figure}
\begin{centering}
\includegraphics[width=\columnwidth]{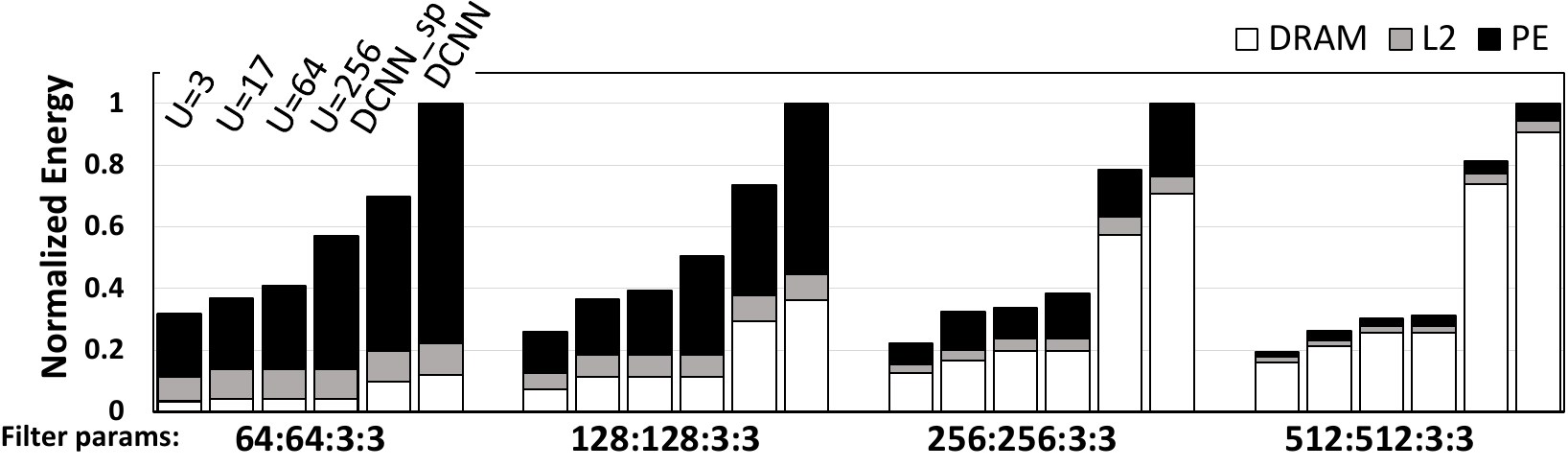}
\caption{\label{eval:energy_breakdown}
       \small 
       Energy breakdown for the 50\% weight density and 16~bit precision point, for specific layers in ResNet.
       Each group of results is for one layer, using the notation $C:K:R:S$.
       All results are relative to \conf{DCNN} for that layer.
       } 
\end{centering}
\end{figure}

To give additional insight, we further break down energy by network layer.
Figure~\ref{eval:energy_breakdown} shows select layers in ResNet-50 given 50\% weight density and 16~bit precision.
Generally, early layers for the three networks (only ResNet shown) have smaller $C$ and $K$; later layers have larger $C$ and $K$.
DRAM access count is proportional to total filter size $R*S*C*K$, making early and later layers compute and memory bound, respectively.
Thus, UCNN reduces energy in early layers by improving arithmetic efficiency and reduces energy in later layers by saving DRAM accesses.





\subsection{Performance Analysis}
\label{sec:jiyong}

We now compare the performance of UCNN to DCNN with the help of two studies.
First, we compare performance assuming no load balance issues (e.g., skip entries in indirection tables; Section~\ref{sec:agr_arch}) and assuming a uniform distribution of weights across filters, to demonstrate the benefit of sparse weights.
Second, we compare performance given real INQ~\cite{INQ} data, taking into account all data-dependent effects.
This helps us visualize how a real implementation of UCNN can differ from the ideal implementation.
For all experiments, we assume the hardware parameters in Table~\ref{tab:bc_hardware_params}.

\begin{figure}[!h]
  \begin{centering}
  \includegraphics[width=\columnwidth]{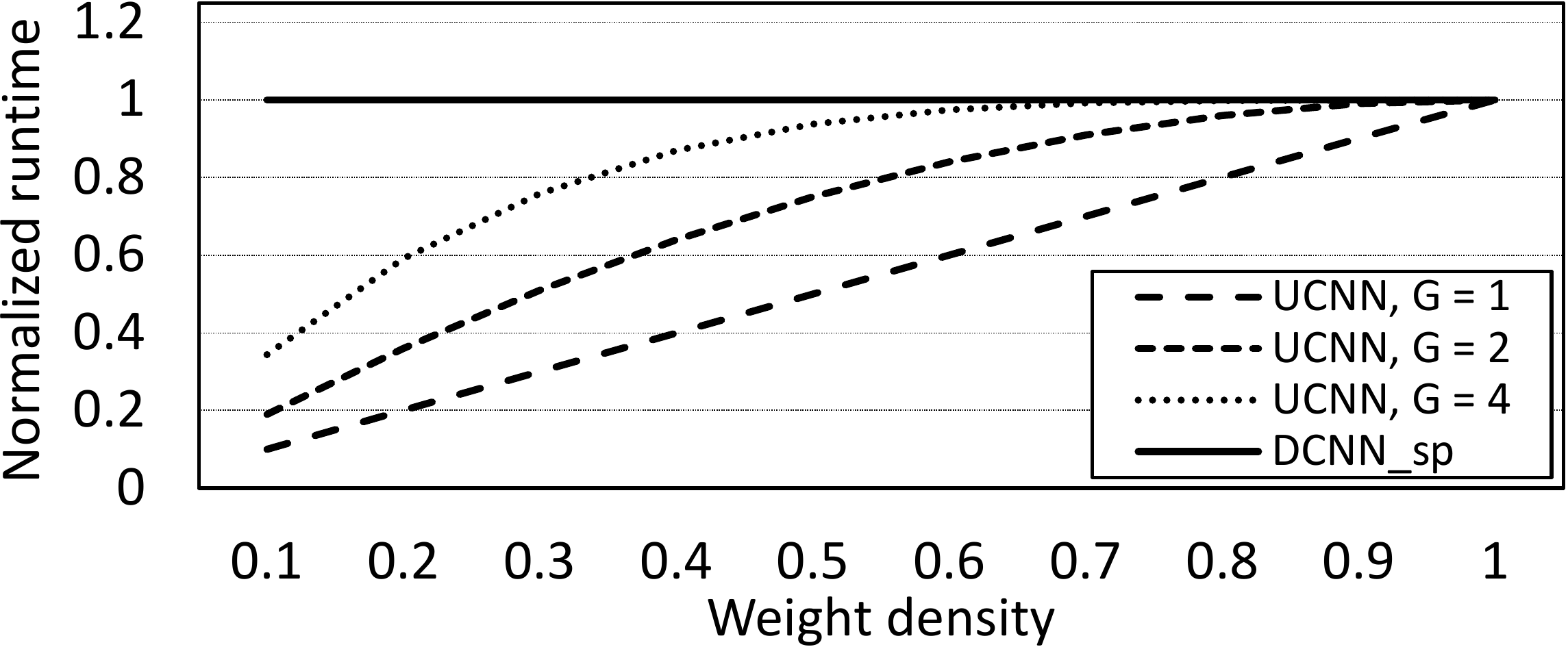}
  \caption{\label{eval:perf_weight_sparse}
           \small 
           Normalized runtime in cycles (lower is better) between \conf{DCNN\_sp} and \conf{UCNN} variants. 
           Runtimes are normalized to \conf{DCNN\_sp}.
           } 
  \end{centering}
\end{figure}

\para{Optimistic performance analysis.}
While all designs in Table~\ref{tab:bc_hardware_params} are throughput-normalized, UCNN can still save cycles due to weight sparsity as shown in Figure~\ref{eval:perf_weight_sparse}.
Potential improvement is a function of $G$: as described in Section~\ref{sec:agr_arch}, the indirection tables with activation group reuse ($G>1$) must store entries corresponding to the union of non-zero weights across the $G$ filters.
This means that choosing $G$ presents a performance energy trade-off: larger $G$ (when this is possible) reduces energy per CNN inference, yet smaller $G$ (e.g., $G=1$) can improve runtime.

\begin{figure*}[h]
  \begin{centering}
  \includegraphics[width=0.9\textwidth]{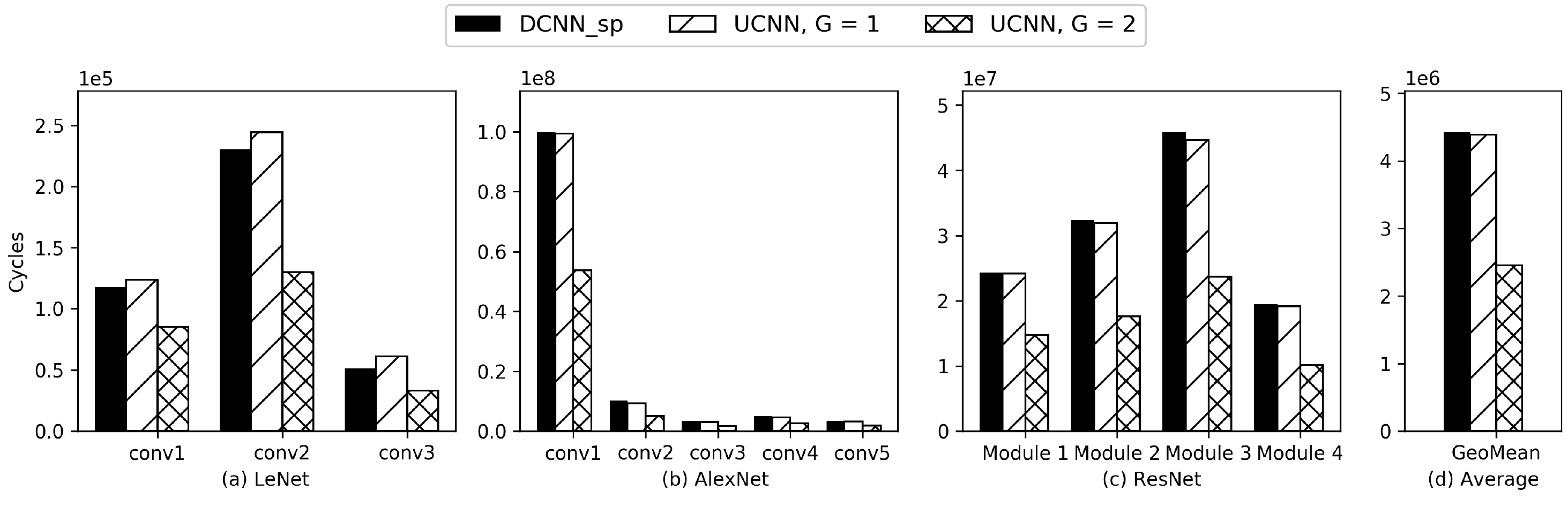}
  \caption{\label{eval:perf_inq_jiyong}
           \small
           Performance study, comparing \conf{DCNN\_sp} ($V_K=1$) and UCNN variants on the three networks from Section~\ref{sec:methods}.
           The geometric means for all variants are shown in (d).
           } 
  \end{centering}
\end{figure*}

\para{Performance on real INQ data.}
We now compare UCNN to DCNN on real INQ~\cite{INQ} training data ($U=17$) and take into account sources of implementation-dependent UCNN performance overhead (e.g., a single multiplier in the PE datapath, and table skip entries; Section~\ref{sec:agr_arch}). The result is presented in Figure~\ref{eval:perf_inq_jiyong}.
Given that our model trained with INQ has 90\% weight density (matching \cite{INQ}), UCNN could improve performance by 10\% in the best case (Section~\ref{sec:eval_fluid}).
However, we see 0.7\% improvement for \conf{UCNN} ($G=1$).
We further observe the following: increasing $V_K=2$ for \conf{DCNN\_sp}, DCNN's performance improves by $2\times$.
However, UCNN $G=2$ (which is throughput-normalized to DCNN $V_K=2$) only improves performance by $1.80\times$, deviating from the ideal improvement of 2$\times$. 
This performance gap is largely due to skip entries in the indirection table (Section~\ref{sec:agr_arch}).
Overall, the performance deficit is dominated by the energy savings with UCNN as presented in Section~\ref{sec:eval_fluid}. Therefore, UCNN still provides a significant performance/watt advantage over DCNN configurations.



\subsection{Model Size (DRAM storage footprint)}
\label{sec:model_size}
\begin{figure}
  \begin{adjustwidth}{-.15cm}{}
  \includegraphics[width=\columnwidth]{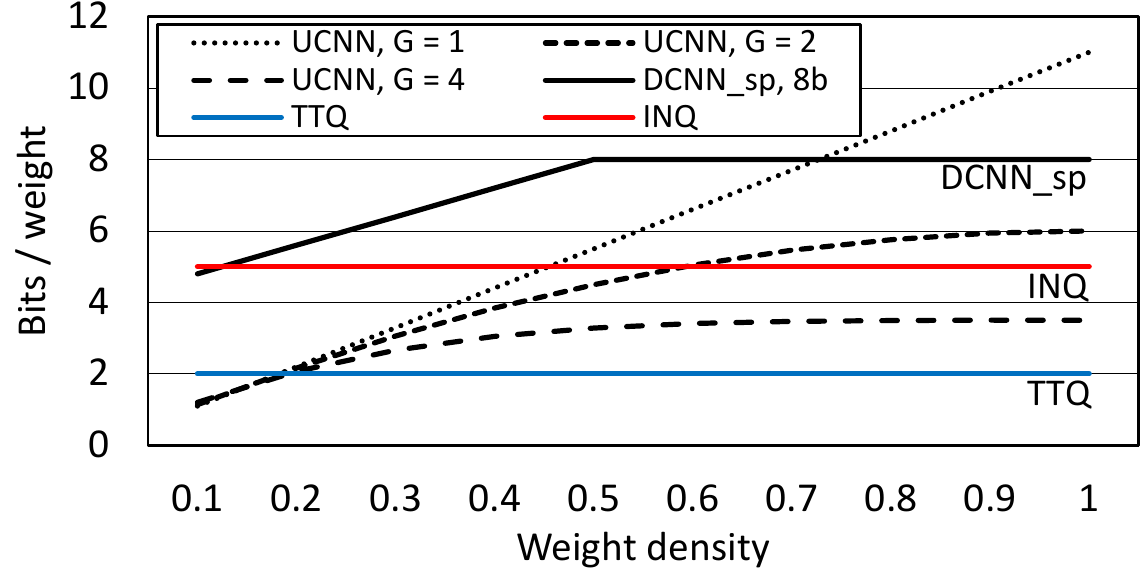}
  \caption{\label{eval:model_size}
           \small
           Model size (normalized per weight), as a function of weight density.
           UCNN indirection table entries are pointers.
           } 
  \end{adjustwidth}
\end{figure}

Figure~\ref{eval:model_size} compares weight compression rates between \conf{UCNN} variants, \conf{DCNN\_sp} and to the stated model sizes in the TTQ~\cite{Ternary} and INQ~\cite{INQ} papers.
UCNN uses activation group reuse and weight sparsity to compress model size (Section~\ref{sec:agr_arch}), however uses the simple pointer scheme from Section~\ref{sec:fdp_arch} to minimize skip entries.
\conf{DCNN\_sp} uses a run-length encoding as discussed in Section~\ref{sec:methods}.
TTQ~\cite{Ternary} and INQ~\cite{INQ} represent weights as 2-bit and 5-bit indirections, respectively.
UCNN, TTQ and INQ model sizes are invariant to the bit-precision per weight. 
This is not true for \conf{DCNN\_sp}, so we only show \conf{DCNN\_sp} with 8~bits per weight to make it more competitive.
TTQ and INQ cannot reduce model size further due to weight sparsity: e.g., a run-length encoding would outweigh the benefit because their representation is smaller than the run-length code metadata.

UCNN models with $G>1$ are significantly smaller than \conf{DCNN\_sp} for all weight densities. However, \conf{UCNN} $G=1$ (no activation group reuse) results in a larger model size than \conf{DCNN\_sp} for models with higher weight density. 



We now compare UCNN's model size with that of TTQ and INQ.
At the 50\% weight density point, \conf{UCNN} $G=4$ (used for TTQ) requires $\sim 3.3$~bits per weight.
If density drops to 30\%, model size drops to $<3$~bits per weight, which \cite{Ternary} shows results in $\sim1\%$ accuracy loss.
At the 90\% weight density point, \conf{UCNN} $G=2$ (used for INQ) requires 5-6 bits per weight.
Overall, we see that UCNN model sizes are competitive with the best known quantization schemes, and simultaneously give the ability to reduce energy on-chip. 

\begin{figure}
  \begin{centering}
  \includegraphics[width=\columnwidth]{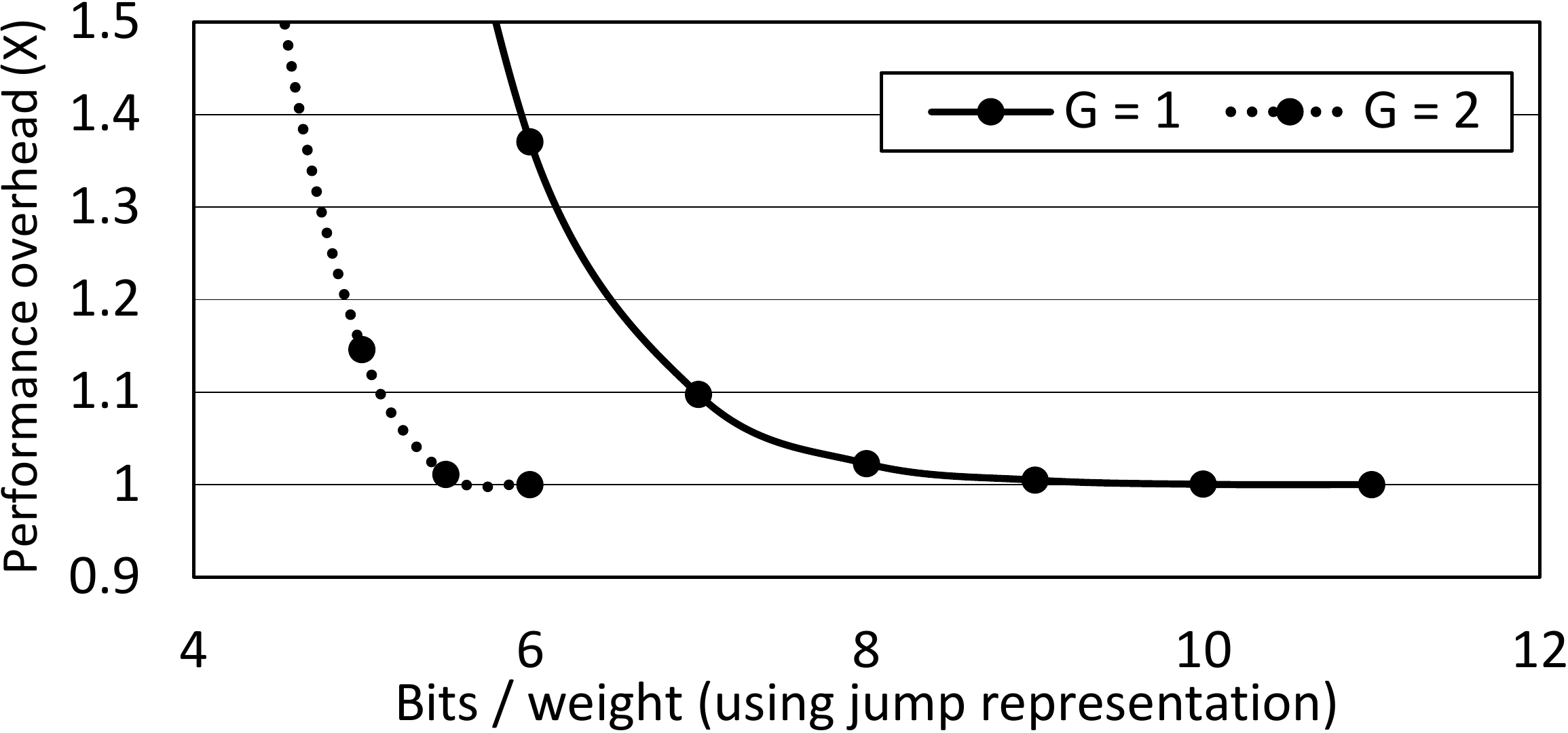}
  \caption{\label{eval:model_size_jt}
          \small 
          UCNN model size (normalized per weight), decreasing jump entry width, for the INQ-trained ResNet.
          } 
  \end{centering}
\end{figure}

\para{Effect of jump-based indirection tables.}
Section~\ref{sec:agr_arch} discussed how to reduce model size for UCNN further by replacing the pointers in the input indirection table with jumps.
The downside of this scheme is possible performance overhead: if the jump width isn't large enough, multiple jumps will be needed to reach the next weight which results in bubbles.
We show these effects on INQ-trained ResNet in Figure~\ref{eval:model_size_jt}.
There are two takeaways.
First, in the $G=1$ case, we can shrink the bits/weight by 3~bits (from 11 to 8) without incurring serious performance overhead ($\sim 2\%$).
In that case, the $G=1$ point never exceeds the model size for \conf{DCNN\_sp} with 8~bit weights.
Second, for the $G=2$ case we can shrink the bits/weight by 1~bit (from 6 to 5), matching INQ's model size with negligible performance penalty.
We note that the same effect can be achieved if the INQ model weight density drops below 60\%.

\subsection{Hardware RTL Results}
\label{sec:kartik}

Finally, Table~\ref{tab:pe_area_breakdown} shows the area overhead of UCNN mechanisms at the PE. 
We implement both DCNN and UCNN PEs in Verilog, using 16~bit precision weights/activations.
Synthesis uses a 32~nm process, and both designs meet timing at 1~GHz.  
Area numbers for SRAM were obtained from CACTI~\cite{cacti} and the area for logic comes from synthesis. 
For a throughput-normalized comparison, and to match the performance study in Section~\ref{sec:jiyong}, we report area numbers for the DCNN PE with $V_K=2$ and the UCNN PE with $G=2, U=17$. 


\begin{table}
{\small
\begin{center}
  \caption{\small
  UCNN PE area breakdown (in $mm^2$).
  }
  \begin{tabular}{l | c | c }
    \hline 
    Component               & DCNN($V_K=2$) & UCNN ($G=2, U=17$)\\
    \hline \hline
    Input buffer            &0.00135      &0.00453\\ 
    Indirection table       & $-$      &0.00100\\
    Weight buffer           &0.00384      &$-$ \\
    Partial Sum buffer      &0.00577      &0.00577 \\
    Arithmetic              &0.00120      &0.00244\\
    Control Logic           &0.00109      &0.00171\\
    \hline
    Total                   &0.01325     &0.01545\\
    \hline 
  \end{tabular}
  \end{center}
}
  \label{tab:pe_area_breakdown}
\end{table} 

Provisioned with a weight buffer $\filt$ of 17 entries, the UCNN PE adds 17\% area overhead compared to a DCNN PE.
If we provision for 256 weights to improve design flexibility (Section~\ref{sec:flexibility}), this overhead increases to 24\%.
Our UCNN design multiplexes a single MAC unit between $G=2$ filters and gates the PE datapath when the indirection table outputs a skip entry (Section~\ref{sec:jiyong}). The RTL evaluation reproduces the performance results from our performance model (Section~\ref{sec:jiyong}).

\section{Related Work}
\label{sec:related}

\para{Weight quantization.}
There is a rich line of work that studies DNN machine efficiency-result accuracy trade-offs by skipping zeros in DNNs and reducing DNN numerical precision (e.g., \cite{DeepCompression,INQ,BNN,Ternary}).
Deep Compression~\cite{DeepCompression}, INQ~\cite{INQ} and TTQ~\cite{Ternary} achieve competitive accuracy on different networks trained on Imagenet~\cite{Imagenet}, although we note that TTQ loses several percent accuracy on ResNet~\cite{Resnet}.
Our work strives to support (and improve efficiency for) all of these schemes in a precision and weight-quantization agnostic fashion.

\para{Sparsity and sparse accelerators.}
DNN sparsity was first recognized by Optimal Brain Damage~\cite{opt_brain_damage} and more recently was adopted for modern networks in Han et al.~\cite{HanPTD15,DeepCompression}.
Since then, DNN accelerators have sought to save cycles and energy by exploiting sparse weights~\cite{Cambriconx}, activations~\cite{Cnvlutin} or both~\cite{SCNN,EIE}.
Relative to our work, these works exploit savings though repeated \emph{zero} weights, whereas we exploit repetition in zero \emph{or non-zero} weights.
As mentioned, we gain additional efficiency through weight sparsity.


\para{Algorithms to exploit computation re-use in convolutions.}
Reducing computation via repeated weights draws inspiration from the Winograd style of convolution~\cite{Winograd}.
Winograd factors out multiplies in convolution (similar to how we factorized dot products) by taking advantage of the predictable filter slide.
Unlike weight repetition, 
Winograd is weight/input ``repetition un-aware'', can't exploit cross-filter weight repetition, loses effectiveness for non-unit strides and only works for convolutions.
Depending on quantization, weight repetition architectures can exploit more opportunity.
On the other hand, Winograd maintains a more regular computation and is thus more suitable for general purpose devices such as GPUs.
Thus, we consider it important future work to study how to combine these techniques to get the best of both worlds.

TTQ~\cite{Ternary} mentions that multiplies can be replaced with a table lookup (code book) indexed by activation.  This is similar to partial produce reuse (Section~\ref{sec:ppr}), however faces challenges in achieving net efficiency improvements.  For example: an 8~bit and 16~bit fixed point multiply in 32~nm is .1 and .4 pJ, respectively.  The corresponding table lookups (512-entry 8~bit and 32K-entry 16~bit SRAMs) cost .17 and 2.5~pJ, respectively~\cite{cacti}.  
Thus, replacing the multiplication with a lookup actually increases energy consumption.  Our proposal gets a net-improvement by reusing compound expressions.

\para{Architectures that exploit repeated weights.}
Deep compression~\cite{DeepCompression} and EIE~\cite{EIE} propose weight sharing (same phenomena as repeated weights) to reduce weight storage, however do not explore ways to reduce/re-use sub computations (Section~\ref{sec:comp_reuse}) through shared weights.
Further, their compression is less aggressive, and doesn't take advantage of overlapped repetitions across filters.




\section{Conclusion}
\label{sec:conclusion}

This paper proposed UCNN, a novel CNN accelerator that exploits weight repetition to reduce on-chip multiplies/memory reads and to compress network model size.
Compared to an Eyeriss-style CNN accelerator baseline, UCNN improves energy efficiency up to $3.7\times$ on three contemporary CNNs.
Our advantage grows to $4\times$ when given dense weights.
Indeed, we view our work as a first step towards generalizing sparse architectures: we should be exploiting repetition in all weights, not just zero weights.

\section{Acknowledgements}

We thank Joel Emer and Angshuman Parasher for many helpful discussions.
We would also like to thank the anonymous reviewers and our shepherd Hadi Esmaeilzadeh, for their valuable feedback.
This work was partially supported by NSF award CCF-1725734.

\bibliographystyle{ieeetr}
\bibliography{bibs/ref}

\begin{thebibliography}{10}

\bibitem{DNN_Speech}
G.~Hinton, L.~Deng, D.~Yu, G.~Dahl, A.~rahman Mohamed, N.~Jaitly, A.~Senior,
  V.~Vanhoucke, P.~Nguyen, B.~Kingsbury, and T.~Sainath, ``Deep neural networks
  for acoustic modeling in speech recognition,'' {\em IEEE Signal Processing
  Magazine}, vol.~29, pp.~82--97, November 2012.

\bibitem{DNN_Image}
D.~Ciregan, U.~Meier, and J.~Schmidhuber, ``Multi-column deep neural networks
  for image classification,'' CVPR'12.

\bibitem{DNN_Economy}
J.~Morajda, ``Neural networks and their economic applications,'' in {\em
  Artificial intelligence and security in computing systems}, pp.~53--62,
  Springer, 2003.

\bibitem{DNN_Medicine}
J.~L. Patel and R.~K. Goyal, ``Applications of artificial neural networks in
  medical science,'' {\em Current clinical pharmacology}, vol.~2, no.~3,
  pp.~217--226, 2007.

\bibitem{DNN_Bio}
H.~Malmgren, M.~Borga, and L.~Niklasson, {\em Artificial Neural Networks in
  Medicine and Biology: Proceedings of the ANNIMAB-1 Conference, G{\"o}teborg,
  Sweden, 13--16 May 2000}.
\newblock Springer Science \& Business Media, 2012.

\bibitem{Alexnet}
A.~Krizhevsky, I.~Sutskever, and G.~E. Hinton, ``Imagenet classification with
  deep convolutional neural networks,'' in {\em Advances in Neural Information
  Processing Systems}, NIPS'12.

\bibitem{TwoStreamConv}
K.~Simonyan and A.~Zisserman, ``Two-stream convolutional networks for action
  recognition in videos,'' NIPS'14.

\bibitem{CNNGan}
A.~Radford, L.~Metz, and S.~Chintala, ``Unsupervised representation learning
  with deep convolutional generative adversarial networks,'' {\em arXiv
  preprint arXiv:1511.06434}, 2015.

\bibitem{DaDianNao}
Y.~Chen, T.~Luo, S.~Liu, S.~Zhang, L.~He, J.~Wang, L.~Li, T.~Chen, Z.~Xu,
  N.~Sun, and O.~Temam, ``Dadiannao: A machine-learning supercomputer,''
  MICRO'14.

\bibitem{Cnvlutin}
J.~Albericio, P.~Judd, T.~Hetherington, T.~Aamodt, N.~E. Jerger, and
  A.~Moshovos, ``Cnvlutin: Ineffectual-neuron-free deep neural network
  computing,'' ISCA'16.

\bibitem{EIE}
S.~Han, X.~Liu, H.~Mao, J.~Pu, A.~Pedram, M.~A. Horowitz, and W.~J. Dally,
  ``{EIE:} efficient inference engine on compressed deep neural network,''
  ISCA'16.

\bibitem{SCNN}
A.~Parashar, M.~Rhu, A.~Mukkara, A.~Puglielli, R.~Venkatesan, B.~Khailany,
  J.~Emer, S.~W. Keckler, and W.~J. Dally, ``Scnn: An accelerator for
  compressed-sparse convolutional neural networks,'' ISCA'17.

\bibitem{TPU}
N.~P. Jouppi, C.~Young, N.~Patil, D.~Patterson, G.~Agrawal, R.~Bajwa, S.~Bates,
  S.~Bhatia, N.~Boden, A.~Borchers, R.~Boyle, P.~Cantin, C.~Chao, C.~Clark,
  J.~Coriell, M.~Daley, M.~Dau, J.~Dean, B.~Gelb, T.~V. Ghaemmaghami,
  R.~Gottipati, W.~Gulland, R.~Hagmann, R.~C. Ho, D.~Hogberg, J.~Hu, R.~Hundt,
  D.~Hurt, J.~Ibarz, A.~Jaffey, A.~Jaworski, A.~Kaplan, H.~Khaitan, A.~Koch,
  N.~Kumar, S.~Lacy, J.~Laudon, J.~Law, D.~Le, C.~Leary, Z.~Liu, K.~Lucke,
  A.~Lundin, G.~MacKean, A.~Maggiore, M.~Mahony, K.~Miller, R.~Nagarajan,
  R.~Narayanaswami, R.~Ni, K.~Nix, T.~Norrie, M.~Omernick, N.~Penukonda,
  A.~Phelps, J.~Ross, A.~Salek, E.~Samadiani, C.~Severn, G.~Sizikov,
  M.~Snelham, J.~Souter, D.~Steinberg, A.~Swing, M.~Tan, G.~Thorson, B.~Tian,
  H.~Toma, E.~Tuttle, V.~Vasudevan, R.~Walter, W.~Wang, E.~Wilcox, and D.~H.
  Yoon, ``In-datacenter performance analysis of a tensor processing unit,''
  ISCA'17.

\bibitem{DeepCompression}
S.~Han, H.~Mao, and W.~J. Dally, ``Deep compression: Compressing deep neural
  network with pruning, trained quantization and huffman coding,'' ICLR'16.

\bibitem{Googlenet}
C.~Szegedy, W.~Liu, Y.~Jia, P.~Sermanet, S.~Reed, D.~Anguelov, D.~Erhan,
  V.~Vanhoucke, and A.~Rabinovich, ``Going deeper with convolutions,'' CVPR'15.

\bibitem{Resnet}
K.~He, X.~Zhang, S.~Ren, and J.~Sun, ``Deep residual learning for image
  recognition,'' CVPR'16.

\bibitem{Ternary}
C.~Zhu, S.~Han, H.~Mao, and W.~J. Dally, ``Trained ternary quantization,'' {\em
  arXiv preprint arXiv:1612.01064}, 2016.

\bibitem{INQ}
A.~Zhou, A.~Yao, Y.~Guo, L.~Xu, and Y.~Chen, ``Incremental network
  quantization: Towards lossless cnns with low-precision weights,'' ICLR'17.

\bibitem{Cambriconx}
S.~Zhang, Z.~Du, L.~Zhang, H.~Lan, S.~Liu, L.~Li, Q.~Guo, T.~Chen, and Y.~Chen,
  ``Cambricon-x: An accelerator for sparse neural networks,'' MICRO'16.

\bibitem{caffecifar}
``Caffe cifar-10 cnn.''
  \url{https://github.com/BVLC/caffe/blob/master/examples/cifar10/cifar10_quick_train_test.prototxt}.

\bibitem{cifar}
A.~Krizhevsky, V.~Nair, and G.~Hinton, ``The cifar-10 dataset,'' {\em online:
  http://www. cs. toronto. edu/kriz/cifar. html}, 2014.

\bibitem{Imagenet}
J.~Deng, W.~Dong, R.~Socher, L.-J. Li, K.~Li, and L.~Fei-Fei, ``{ImageNet: A
  Large-Scale Hierarchical Image Database},'' CVPR'09.

\bibitem{BatchNorm}
S.~Ioffe and C.~Szegedy, ``Batch normalization: Accelerating deep network
  training by reducing internal covariate shift,'' {\em arXiv preprint
  arXiv:1502.03167}, 2015.

\bibitem{Cong2014MinimizingCI}
J.~Cong and B.~Xiao, ``Minimizing computation in convolutional neural
  networks,'' in {\em ICANN}, 2014.

\bibitem{HanPTD15}
S.~Han, J.~Pool, J.~Tran, and W.~J. Dally, ``Learning both weights and
  connections for efficient neural networks,'' NIPS'15.

\bibitem{precision}
V.~Vanhoucke, A.~Senior, and M.~Z. Mao, ``Improving the speed of neural
  networks on cpus,'' in {\em Proc. Deep Learning and Unsupervised Feature
  Learning NIPS Workshop}, vol.~1, p.~4, Citeseer, 2011.

\bibitem{Eyeriss}
Y.-H. Chen, J.~Emer, and V.~Sze, ``Eyeriss: A spatial architecture for
  energy-efficient dataflow for convolutional neural networks,'' ISCA'16.

\bibitem{Relu}
A.~L. Maas, A.~Y. Hannun, and A.~Y. Ng, ``Rectifier nonlinearities improve
  neural network acoustic models,'' ICML'13.

\bibitem{marktalk}
M.~Horowitz, ``Computing's energy problem (and what we can do about it).''
  ISSCC, 2014.

\bibitem{cacti}
N.~Muralimanohar and R.~Balasubramonian, ``Cacti 6.0: A tool to understand
  large caches,'' 2009.

\bibitem{cacti-lowswing}
A.~N. Udipi, N.~Muralimanohar, and R.~Balasubramonian HiPC'09.

\bibitem{BNN}
M.~Courbariaux and Y.~Bengio, ``Binarynet: Training deep neural networks with
  weights and activations constrained to +1 or -1,'' NIPS'16.

\bibitem{opt_brain_damage}
Y.~LeCun, J.~S. Denker, and S.~A. Solla, ``Optimal brain damage,'' in {\em
  Advances in Neural Information Processing Systems 2} (D.~S. Touretzky, ed.),
  pp.~598--605, Morgan-Kaufmann, 1990.

\bibitem{Winograd}
A.~Lavin and S.~Gray, ``Fast algorithms for convolutional neural networks,''
  CVPR'16.

\end{thebibliography}

\end{document}